\documentclass[english, twoside, 11pt]{article}

\usepackage{jmlr2e}

\usepackage{babel}
\usepackage{subfigure}

\usepackage{color}
\graphicspath{{./img/}}
\definecolor{red}{RGB}{204,0,51}
\definecolor{orange}{RGB}{255,102,0}
\definecolor{lightgreen}{RGB}{102,153,0}
\definecolor{green}{RGB}{0,153,0}
\definecolor{blue}{RGB}{0,51,255}
\definecolor{purple}{RGB}{102,0,255}

 \ifx\proof\undefined\
   \newenvironment{proof}[1][\proofname]{\par
     \normalfont\topsep6\p@\@plus6\p@\relax
     \trivlist
     \itemindent\parindent
     \item[\hskip\labelsep
           \scshape
       #1]\ignorespaces
   }{%
     \endtrivlist\@endpefalse
   }
   \providecommand{\proofname}{Proof}
 \fi

\newcommand{\algblockone}[8]{
	\begin{tabular}{l}
		\Large $\sigma = #1$ \\
		\hline
		\begin{tabular}{lll}
			\footnotesize \textcolor{red}{\rule{2mm}{2mm}} $\epsilon$-greedy, $\epsilon = #2$ (#3) &
			\footnotesize \textcolor{lightgreen}{\rule{2mm}{2mm}} Pursuit, $\beta = #6$ (#7) &
			\footnotesize \textcolor{blue}{\rule{2mm}{2mm}} UCB1 (#8) \\
			\footnotesize \textcolor{orange}{\rule{2mm}{2mm}} Softmax, $\tau = #4$ (#5) &
}
\newcommand{\algblocktwo}[6]{
			\footnotesize \textcolor{green}{\rule{2mm}{2mm}} Reinforcement comparison, $\alpha = #1, \beta = #2$ (#3) &
			\footnotesize \textcolor{purple}{\rule{2mm}{2mm}} UCB1-Tuned (#4) \\
		\end{tabular}\\
		\subfigure[Regret per turn]{\includegraphics[width=7cm]{#5}}
		\subfigure[Percentage of optimal arm plays]{\includegraphics[width=7cm]{#6}}
	\end{tabular} \\
}

\jmlrheading{1}{2000}{1-48}{4/00}{10/00}{Volodymyr Kuleshov and Doina Precup}

\ShortHeadings{Algorithms for the multi-armed bandit problem}{Kuleshov and Precup}
\firstpageno{1}

\begin{document}

\title{Algorithms for the multi-armed bandit problem}

\author{\name Volodymyr Kuleshov \email volodymyr.kuleshov@mail.mcgill.ca \\
	\name Doina Precup \email dprecup@cs.mcgill.ca\\
       \addr School of Computer Science\\
       McGill University}

\editor{Leslie Pack Kaelbling}

\maketitle

\begin{abstract}
The stochastic multi-armed bandit problem is an important model for studying the exploration-exploitation tradeoff in reinforcement learning.
Although many algorithms for the problem are well-understood theoretically, empirical confirmation of their effectiveness is generally scarce.
This paper presents a thorough empirical study of the most popular multi-armed bandit algorithms. Three important observations can be made from our results. 
Firstly, simple heuristics such as $\epsilon$-greedy and Boltzmann exploration outperform theoretically sound algorithms on most settings by a significant margin. 
Secondly, the performance of most algorithms varies dramatically with the parameters of the bandit problem. Our study identifies for each algorithm the settings where it performs well, and the settings where it performs poorly. These properties are not described by current theory, even though they can be exploited in practice in the design of heuristics. 
Thirdly, the algorithms' performance relative each to other is affected only by the number of bandit arms and the variance of the rewards. This finding may guide the design of subsequent empirical evaluations.

In the second part of the paper, we turn our attention to an important area of application of bandit algorithms:
clinical trials. Although the design of clinical trials has been one of the principal practical problems motivating research
on multi-armed bandits, bandit algorithms have never been evaluated as potential treatment allocation strategies.
Using data from a real study, we simulate the outcome that a 2001-2002 clinical trial would have had if bandit algorithms had been used to allocate patients to treatments.
We find that an adaptive trial would have successfully treated at least 50\% more patients,
while significantly reducing the number of adverse effects and increasing patient retention. 
At the end of the trial, the best treatment could have still been identified with a high level of statistical confidence.
Our findings demonstrate that bandit algorithms are attractive alternatives to current adaptive treatment allocation strategies.

\end{abstract}


\section{Introduction}

Multi-armed bandit problems have been introduced by Robbins (1952)
and have since been used extensively to model the trade-offs faced
by an automated agent which aims to gain new knowledge by exploring
its environment and to exploit its current, reliable knowledge.
Such problems arise frequently in practice, for example in the context
of clinical trials or on-line advertising. The multi-armed bandit
problem offers a very clean, simple theoretical formulation for analyzing
trade-offs between exploration and exploitation. A comprehensive overview
of bandit problems from a statistical perspective is given in Berry
\& Fristedt (1985)\nocite{berry85}.

In its simplest formulation (generally referred to as \emph{stochastic}),
a bandit problem consists of a set of K probability distributions
$\langle D_1, \dots, D_K\rangle$ with associated expected values $\langle \mu_1, \dots, \mu_K\rangle$ and variances 
$\langle\sigma_1^2, \dots, \sigma_k^2\rangle$. Initially, the $D_i$ are unknown to the player.  In fact, these distributions are generally interpreted as corresponding to arms on a slot machine; the player is viewed as a gambler whose goal is to collect as much money as possible by pulling these arms over many turns. 
At each turn, $t = 1,2,...$, the player selects an arm, with index $j(t)$, and receives a reward $r(t) \sim D_{j(t)}$.  
The player has a two-fold goal: on one hand, finding out which distribution has the highest expected value; on 
the other hand, gaining as much rewards as possible while playing.  Bandit algorithms specify a strategy by
which the player should choose an arm $j(t)$ at each turn.

The most popular performance measure for bandit algorithms is the \emph{total expected regret}, defined for any fixed turn $T$ as:
\[
R_T =T\mu^*-\sum_{t=1}^T \mu_{j(t)}
\]
where $\mu^*=\max_{i=1, \dots k} \mu_i$ is the expected reward from the best arm. 

Alternatively, we can express the total expected regret as 
\[
R_T =T\mu^*-\mu_{j(t)}\sum_{k=1}^K \mathbb{E}(T_k(T))
\]
where $T_k(T)$ is a random variable denoting the number of plays of arm $k$ during the first $T$ turns.

A classical result of Lai and Robbins (1985) states that for any suboptimal arm $k$,
\[
\mathbb{E}(T_k(T)) \geq \frac{\ln T}{D(p_k || p^*)}
\]
where $D(p_j || p^*)$ is the Kullback-Leibler divergence between the reward density $p_k$ of the suboptimal arm and the reward density $p^*$ of the optimal arm, defined formally as
\[
D(p_k || p^*) =  \int p_j  \ln \frac{p_j}{p^*}
\]

Regret thus grows at least logarithmically, or more formally, $R_T = \Omega (\log T)$. An algorithm is said to {solve} the multi-armed bandit problem if it can match this lower bound, that is if $R_T = O(\log T)$.

\subsection*{Evaluating algorithms for the bandit problem}
Many theoretical bounds have been 
established for the regret of different bandit algorithms in recent years (e.g., Auer et al., 2002\nocite{auer02}, Audibert et al., 2009\nocite{audibert09}, Cesa-Bianchi and Fisher, 1998\nocite{cesabianchi98}, Rajaraman \& Sastry, 1996\nocite{rajaraman96}).  However, theoretical analyses are not available for all algorithms, and existing bounds are generally too loose to accurately measure the strategies' performance.

Empirical evidence regarding the performance of many algorithms is unfortunately also limited. Algorithms are most often evaluated within a larger body of mostly theoretical work. Evaluation done in this context is often performed on a small number of bandit problem instances (for example, on bandits with small numbers of arms) that may not generalize to other settings. Moreover, different authors evaluate their algorithms in different settings, which complicates the comparison of algorithms.

The most extensive empirical study so far that compares multiple algorithms has been done by Vermorel and Mohri (2005)\nocite{vermorel05}.  This paper does not include an evaluation of the UCB family
of algorithms, which has recently become very popular, as well as that of some other important strategies, such as pursuit or reinforcement comparison (see Sutton and Barto, 1998). Moreover, the paper does not investigate the effect of the variance
of the rewards on algorithm performance. In our experiments, that effect has turned out to be very significant.
The paper also does not attempt to tune the algorithms optimally; some algorithms may therefore be underperforming. Finally, the paper not describe in full detail the criteria used to evaluate performance, making it difficult to interpret the results.

In this paper, we address this gap in the bandit literature by providing an extensive empirical evaluation of the most popular bandit strategies.
We conduct a thorough analysis to identify the aspects of a bandit problem that affect the performance of algorithms relative to each other. Surprisingly, the only relevant characteristics turn out to be number of arms and the reward distributions' variance.

We proceed to measure algorithm performance under different combinations of these two parameters. Remarkably, the simplest heuristics outperform more sophisticated and theoretically sound algorithms on most settings. Although a similar observation has been made by Vermorel and Mohri (2005), our experiments are the first that strongly suggest that this occurs on practically every bandit problem instance.

We also observe that the performance of any algorithm varies dramatically from one bandit problem instance to the other.  We identify for each algorithm the settings where it performs well, and the settings where it performs poorly. Such properties are not described by current theory, even though they can be exploited within heuristics for solving real-world problems. 

As an important side result, our study precisely identifies the aspects of a bandit problem that must be considered in an experiment. They are the number of arms and the reward variance. We hope this observation will be taken into account in the design of subsequent empirical studies.

From a theoretical viewpoint, our findings indicate the need for a formal analysis of simple heuristics, and more generally for the development of theoretically sound algorithms that perform as well as the simpler heuristics in practice.

\subsection*{Clinical trials from a multi-armed bandit viewpoint}

In the second half of the paper, we turn our attention to an important application of bandit algorithms: clinical trials. The design of clinical trials is one of the main practical problems that motivates research on multi-armed bandits, and several seminar papers in the field (Robbins, 1952, Gittins, 1989)\nocite{rob52,gittins89} describe it as such\footnote{Gittins and Jones (1979)\nocite{GJ79}, for example, refer to clinical trials as the ``chief practical motivation [for the design of bandit algorithms]''.}.
Indeed, a clinical trial perfectly captures the problem of balancing exploration and exploitation: we are looking for a way to simultaneously identify the best treatment
(the best ``arm'') and ensure that as much patients as possible
can benefit from it. 

In a traditional clinical trial, patients are randomized into
two equal-sized groups. The best treatment can usually be identified with a high level of confidence, but only half of the patients benefit from it. Adaptive trials that dynamically allocate more patients
to the better treatment have long been advocated
for ethical reasons, but even after decades of theoretical discussion, their use remains very limited (Chow and Chang, 2008\nocite{CC08}). 

Modern adaptive clinical trials can be classified into several families. Group sequential designs are trials that can be stopped prematurely based on interim results, such as the performance of a particular treatment. Sample size re-estimation designs allow the patient population size to be readjusted in the course of the trial. Drop-the-losers designs, on the other hand, allow certain treatments to be dropped or added. Naturally, such trials drop less promising treatments first. Other types of adaptive trials include adaptive dose finding designs, adaptive treatment-switching designs, as well as multiple adaptive designs, which combine features from one or more families described above. For a thorough discussion of the literature on adaptive clinical trials, see the survey paper by Chow and Chang (2008)\nocite{CC08}.

In our context, the most interesting family of adaptive trials are so-called adaptive randomization designs. Such designs adjust the patients' treatment assignment probabilities in favor of more successful treatments during the course of the trial. One of the most popular adaptive randomization strategies is \emph{play-the-winner}, which operates similarly to the pursuit family of bandit algorithms. Due to their simplicity, adaptive randomization strategies form arguably the most popular family of adaptive trials. However, randomization strategies are often based on ad-hoc heuristics that offer few, if any, guarantees on patient welfare. In terms of formal guarantees, bandit algorithms have an advantage over strategies like play-the-winner.

Even though there is an extensive literature on adaptive clinical trials, to our knowledge there is no study that evaluates bandit algorithms as treatment allocation strategies. This is particularly surprising given the fact that many bandit algorithms were specifically designed for that purpose.

Moreover, clinical trial simulations are seldom based on real clinical data (Orloff et al., 2009\nocite{orloff2009}). Yao and Wei (1996)\nocite{YW96} conducted one of the only simulation studies of which we are aware that is based on data from an actual clinical trial. They demonstrate that if the \emph{play-the-winner} adaptive randomization strategy had been used in the 1992 trial of the HIV drug zidovudine, significantly more patients could have been successfully treated, and the effectiveness of the drug could have still been established with high confidence.


\subsection*{Applying bandit algorithms in clinical trials}

In this work, our aim is two fold. In the spirit of Yao and Wei (1996)\nocite{YW96}, we want to determine whether bandit algorithms constitute feasible and effective adaptive trial strategies, and more generally we wish to produce an evaluation of the effectiveness of adaptive clinical trials based on real-world data.

In particular, we aim to answer the following three
questions:
\begin{enumerate}
\item Is it feasible to implement bandit strategies given real world constraints
such as patients' arrival dates, a long treatment time,
patient dropout, etc.?
\item At the end of the trial, can we identify the
best treatment with a good level of statistical confidence (with a
small p-value)?
\item Do bandit-based adaptive trials offer a significant advantage over traditional
trials in terms of patient welfare?
\end{enumerate}
A thorough overview of the literature on simulating clinical trials can be found in Holford et al. (2000)\nocite{HKMP00}.

To answer these questions, we simulate using real data what would have happened if a 2001-2002 clinical trial of opioid addiction treatments used adaptive bandit strategies instead of simple randomization. To measure the effectiveness of each approach, we use a variety of criteria, including
the number of patients successfully treated, patient retention, the number of adverse
effects, and others.

We find that bandit-based treatments would have allowed at least 50\% more patients to be successfully treated,
while significantly reducing the number of adverse effects and increasing patient retention. 
At the end of the trial, the best treatment could have still been identified with a high level of statistical confidence.
Our findings demonstrate that bandit algorithms are attractive alternatives to current treatment allocation strategies.

\subsection*{Outline}

The paper is organized as follows.  In Section \ref{sec:algs}, we briefly review the studied algorithms.  Section \ref{sec:setup}
describes the setup used for the experiments.  Section \ref{sec:results} presents a selection of representative results.  In Section \ref{sec:disc} we discuss three main conclusions that can be drawn from our results as well as their implications for subsequent theoretical and empirical work.

In Section \ref{sec:trials_intro}, we briefly resummarize our motivation for considering clinical trials in the context of multi-armed bandit algorithms. In Section \ref{sec:trials_desc}, we present the trial on which we base our simulation, and in Section \ref{sec:trials_setup} we describe in detail how that simulation is performed. In Section \ref{sec:trials_results}, we present our results, and in Section \ref{sec:trials_disc} we discuss the three questions our simulation was intending to answer in light of the results obtained. We conclude in Section~\ref{sec:concl}.


\section{Algorithms}\label{sec:algs}

In this section, we briefly present the six algorithms that will be evaluated throughout the paper along with their known theoretical properties. 

The first four algorithms are $\epsilon$-greedy, Boltzmann exploration, pursuit, and reinforcement comparison. Each of these heuristics captures distinct ideas on handling the exploration/exploitation tradeoff (Sutton and Barto, 1998). For example, Boltzmann exploration is based on the principle that the frequency of plays of an arm should be proportional to its average reward; the key idea of pursuit is to maintain an explicit probability distribution over the arms, and to search directly in the space of probability distributions. Even though each idea represents an important approach to handling the exploration/exploitation tradeoff, none of the heuristics is well understood theoretically. We are not even aware of any empirical study that systematically evaluates pursuit and reinforcement comparison. 

In contrast, the latter two algorithms, UCB1 and UCB1-Tuned, are based on sophisticated mathematical ideas that are exploited to provide strong theoretical guarantees on the expected regret. In fact, the UCB1 algorithm solves the multi-armed bandit problem optimally up to a constant factor in a way that will be made precise below. How the former intuitive heuristics compare to the more sophisticated algorithms is an important question to which the answer is not clearly known.

In order to define the algorithms, we will use the following notation. Most strategies 
maintain empirical reward means for each arm that are updated at every turn $t$.  We denote by $\hat{\mu}_i(t)$ the empirical mean of arm $i$ after $t$ turns.  The probability of picking arm $i$ at time $t$ will be denoted $p_i(t)$.


\subsection*{$\epsilon$-greedy}

The $\epsilon$-greedy algorithm is widely used because it is very simple, and has obvious generalizations for sequential 
decision problems.  At each round $t = 1,2,...$ the algorithm selects the arm with the highest empirical mean with probability $1-\epsilon$, and selects a random arm with probability $\epsilon$.  In other words, given initial empirical means $\hat{\mu}_1(0),...,\hat{\mu}_K(0)$,
\[
p_i(t+1) = \left\{\begin{array}{ll} 1-\epsilon+\epsilon/k & \mbox{ if } i=\arg\max_{j=1,...,K} \hat{\mu}_j(t) \\ \epsilon/k & \mbox{otherwise.}\end{array}\right.
\]
If $\epsilon$ is held constant, only a linear bound on the expected regret can be achieved. 
Cesa-Bianchi and Fisher (1998)\nocite{cesabianchi98} proved poly-logarithmic bounds for variants of the algorithm in which $\epsilon$ decreases with time. In an earlier empirical study, Vermorel and Mohri (2005)\nocite{vermorel05} did not find any practical advantage to using these methods. Therefore, in our experiments, we will only consider fixed values of $\epsilon$.


\subsection*{Boltzmann Exploration (Softmax)}

Softmax methods are based on Luce's axiom of choice (1959)\nocite{luce59} 
and pick each arm with a probability that is proportional to its average reward. 
Arms with greater empirical means are therefore picked with higher probability.
In our experiments, we study Boltzmann exploration (see, e.g., Sutton \& Barto, 1998), a Softmax method that selects an arm using a Boltzmann distribution. Given initial empirical means $\hat{\mu}_1(0),...,\hat{\mu}_K(0)$,
\[
p_i(t+1) = \frac{e^{\hat{\mu}_i(t)/\tau}}{\sum_{j=1}^k e^{{\hat{\mu}_j(t)}/\tau}}, i=1\dots n
\]
where $\tau$ is a temperature parameter, controlling the randomness of the choice. When $\tau = 0$, Boltzmann Exploration acts like pure greedy. As $\tau$ tends to infinity, the algorithms picks arms uniformly at random. 

As in the case of $\epsilon$-greedy, polylogarithmic regret bounds exist \cite{cesabianchi98} for decreasing $\tau$ schedules.  However, because empirical evidence suggests that such schedules offer no practical advantage \cite{vermorel05}, we will use fixed values of $\tau$ in our experiments.


\subsection*{Pursuit Algorithms}

The methods explained so far are essentially based on the estimated value of each arm.  In contrast, pursuit algorithms (Thathachar \& Sastry, 1985)\nocite{thathachar85} maintain an explicit policy over the arms, whose updates are informed by the empirical means, but are performed separately.  We use the simple version of the pursuit algorithm described in \cite{sutton98}.  The algorithms starts with uniform probabilities assigned to each arm, $p_i(0)=1/k$.  At each turn $t$, the probabilities are re-computed as follows:
\[
p_i(t+1) = \left\{\begin{array}{ll} p_i(t) + \beta(1-p_i(t)) & \mbox{ if }i=\arg\max_j \hat{\mu}_j(t)\\ p_i(t)+ \beta(0-p_i(t)) & \mbox{ otherwise }\end{array}\right.
\]
where $\beta\in(0,1)$ is a learning rate.
Pursuit algorithms are related to actor-critic algorithms used in reinforcement learning for sequential decision problems.
Rajaraman and Sastry (1996)\nocite{rajaraman96} provide PAC-style convergence rates for different forms of the pursuit algorithm, in the context of learning automata.


\subsection*{Reinforcement Comparison}

Reinforcement comparison methods \cite{sutton98} are similar to pursuit methods, in that they maintain a distribution over actions which is not computed directly from the empirical means. These methods also maintain an average expected reward $\bar{r}(t)$. The probability of selecting an arm is computed by comparing its empirical mean with $\bar{r}(t)$. The probability  will be increased if it is above average, and decreased otherwise.  Intuitively, this scheme is designed to account for cases in which arms have very similar value. 

More formally, the algorithm maintains a set of preferences, $\pi_i(t)$, for each arm $i$.  At each turn $t=1,2,...$, the probability $p_i(t)$ is computed using a Boltzmann distribution based on these preferences:
\[
p_i(t) = \frac{e^{\pi_i(t)}}{\sum_{j=1}^k e^{\pi_j(t)}}
\]
If arm $j(t)$ is played at turn $t$, and reward $r(t)$ is received, the preference $\pi_{j(t)}$ is updated as:
\[
\pi_{j(t)}(t+1) = \pi_{j(t)}(t) + \beta(r(t)-\bar{r}(t))
\]
Also, at every turn, the mean of the rewards is updated as:
\[
\bar{r}(t+1) = (1-\alpha) \bar{r}(t) + \alpha r(t)
\]
Here, $\alpha$ and $\beta$ are learning rates between $0$ and $1$.
To our knowledge, no theoretical analysis of the regret of reinforcement comparison methods exists to date.


\subsection*{Upper Confidence Bounds (UCB)}

The UCB family of algorithms has been proposed by Auer, Cesa-Bianchi \& Fisher (2002)\nocite{auer02} as a simpler, more elegant implementation of the idea of optimism in the face of uncertainty, proposed by by Lai \& Robbins (1985)\nocite{lai85}. 
An extension of UCB-style algorithms to sequential, tree-based planning was developed by Kocsis \& Szepesvari (2006)\nocite{kocsis06}, and it has proven very successful in Go playing programs.

The simplest algorithm, UCB1, maintains the number of times that each arm has been played, denoted by $n_i(t)$, in addition to the empirical means.  Initially, each arm is played once.  Afterwards, at round $t$, the algorithm greedily picks the arm $j(t)$ as follows:
\[
j(t) = \arg\max_{i=1\dots k} \left(\hat{\mu}_i + \sqrt{\frac{2\ln t}{n_i}}\right)
\]
Auer, Cesa-Bianchi \& Fisher (2002) show that at turn $t$, the expected regret of UCB1 is bounded by:
\[
8\sum_{i: \mu_i<\mu^*} \frac{\ln t}{\Delta_i} + \left(1+ \frac{\pi^2}{3}\right)\sum_{i=1}^k \Delta_i
\]
where $\Delta_i=\mu^*-\mu_i$. This $O(\log n)$ bound on the regret matches a well-known $\Omega(\log n)$ bound by Lai and Robbisn (1989?). Hence UCB1 achieves the optimal regret up to a multiplicative constant, and is said to \emph{solve} the multi-armed bandit problem.

The authors also propose another algorithm, UCB1-Tuned, which they claim performs better in practice but comes without theoretical guarantees. The main feature of UCB1-Tuned is that it takes into account the variance of each arm and not only its empirical mean.  More specifically, at turn $t=1,2,...$ the algorithm picks an arm $j(t)$ as
\[
j(t) =  \arg\max_{i=1\dots k} \left(\hat{\mu}_i + \sqrt{\frac{\ln t}{n_i}\min{\left(\frac{1}{4},V_i(n_i)\right)}} \right)
\]
where 
\[V_i(t)=\hat{\sigma}^2_i(t) + \sqrt{\frac{2 \ln t}{n_i(t)}}.
\]  The estimate of the variance $\hat{\sigma}^2_i(t)$ can be computed as usual by maintaining the empirical sum of squares of the reward, in addition to the empirical mean.
Audibert, Munos \& Szepesvari (2009)\nocite{audibert09} provide expected regret bounds and regret concentration results for variance-based UCB algorithms similar to UCB1-Tuned.


\section{Experimental Setup}\label{sec:setup}


An instance of the bandit problem is fully characterized by the number of arms and the arms' reward distributions. However, not every aspect of these distributions affects the relative performance of the algorithms. Clearly, the number of arms and the reward variance will affect performance. Quite surprisingly, they will turn out to be the only characteristics of a bandit that need to be considered. Moments higher than the variance are of little importance when comparing algorithms. 

The goals of our experimental setup are thus twofold: identify the characteristics of a bandit problem that affect algorithm performance, and evaluate how exactly they affect that performance. To achieve that, we will vary in isolation the number of arms and their variance, as well as the type of the reward distribution (to account for the higher moments) and the distribution of the expected values of the arms on $[0,1]$. 
Most algorithms admit parameters and we tune each algorithm optimally for each experiment.

In this section, we will describe in detail how all of this is done. 

\subsection*{Overall setup}

Each experiment runs for 1000 turns, as this is the time by which all learning curves plateau. 

We report on three performance criteria:
\begin{enumerate}
\item
The total regret accumulated over the experiment.
\item
The regret as a function of time.
\item
The percentage of plays in which the optimal arm is pulled.
\end{enumerate}
The first criterion summarizes the performance of the algorithm in one number, while the second illustrates in more detail how the algorithm handled the problem. The third criterion is relevant for situations in which minimizing the number of suboptimal arm plays is important (a clinical trial would be an obvious example).

Every experiment was repeated 1000 times, and at every repetition, the expected values of the $K$ reward distributions were chosen uniformly at random on $[0,1]$. Results were averaged over these 1000 independent runs.


\subsection*{Number of arms and variance}

We evaluate the algorithms on settings where the number of arms $K$ equals $2, 5, 10, 50$. The value $K=2$ is only included because this is a special case with some practical applications.  Values of $K$ larger than $50$ provide tasks that are difficult for all algorithms.  However, their relative behavior is consistent with the $K=50$ case even at higher values. We find that the values $K=5, 10$ represent good benchmarks for settings with small and medium numbers of arms, respectively.

Unless indicated otherwise, rewards are sampled from a normal distribution. We evaluate the algorithms on settings where the variance parameter $\sigma^2$ equals $0.01^2, 0.1^2, 1^2$ for every arm. These values correspond to standard deviations that are 1\%, 10\%, and 100\% of the interval [0,1] containing the expected values.  Obviously, smaller variance leads to an easier problem, as the arms are well separated. 


\subsection*{Other bandit characteristics}

To evaluate the effects of higher moments, we experimented with several types of reward distributions (normal, uniform, triangular, inverse Gaussian and Gumbel). 
In each case, the parameters of the distribution were chosen to obtain identical expected values and variances. Results were very similar for all distributions, and this is why we use only the normal distribution in our main set of experiments. 

The means of the arms were randomly chosen on the interval $[0,1]$ at each of the 1000 repetitions.  We experimented with choosing the means both according to a uniform distribution on $[0,1]$, and according to a normal distribution ${\cal N}(0.5, \sqrt{12})$ ($\sqrt{12}$ is the variance of the uniform distribution).    The results were very similar in terms of the ranking and behavior of the algorithms, and for our main experiments, we report only the results that use the uniform distribution. 

For the algorithms  that require initial values of empirical means, we always use optimistic initialization and start with initial values of $1$.  We found that this choice always results in the best performance.


\subsection*{Parameter tuning}

On every experiment, the algorithms were tuned for maximum performance 
according to the first criterion (since this is the usual criterion used in the literature). This almost always leads to the best possible performance on the second criterion and good performance on the third.  These optimized parameter settings are included in the legends, next to the graphs.


\section{Empirical Results}\label{sec:results}

In this section, we first present our main set of results, obtained from evaluating the algorithms on twelve multi-armed bandit instances. Each instance is characterized by a different combination of $K$ and $\sigma^2$, the number of arms and the reward variance.

In the second half of the section, we present results demonstrating that other aspects of the bandit (such as the higher moments of the reward distribution) are not important for the purpose of evaluating algorithm performance. Finally, while tuning the algorithms we have observed that parameters dramatically affect performance, and we include material that illustrates that.

\subsection*{Number of arms and variance}

\begin{figure*}[hp]

\algblockone{0.01}{0.005}{1.21}{0.001}{0.351}{0.1}{55.5}{19.8}
\algblocktwo{0.4}{0.98}{3.64}{5.26}{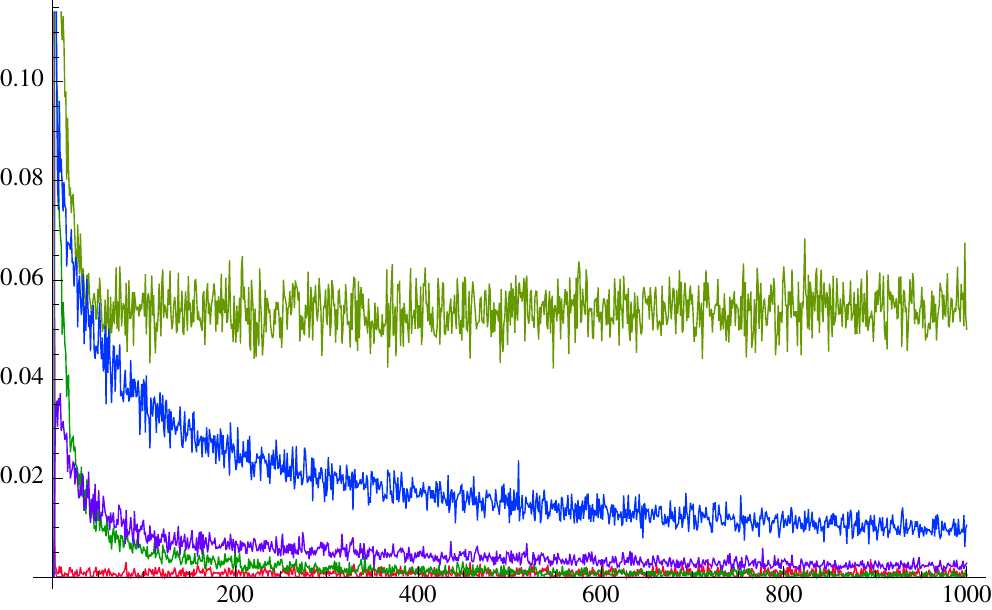}{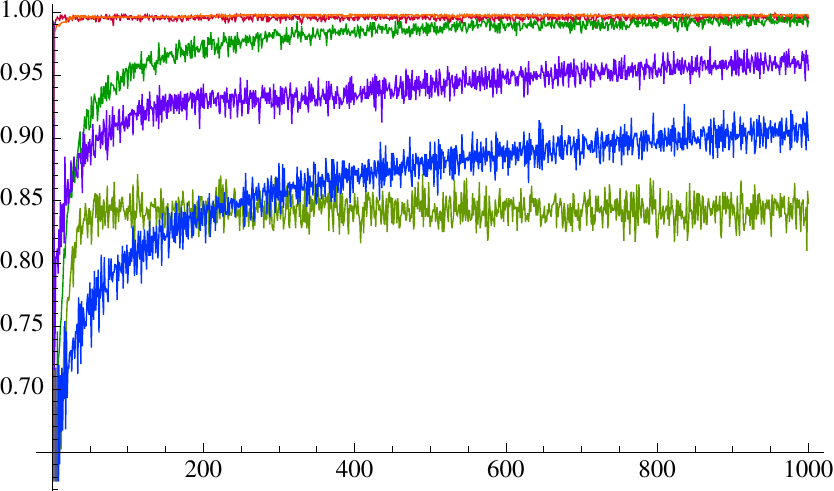}

\algblockone{0.1}{0.001}{3.15}{0.01}{1.53}{0.1}{55.3}{19.8}
\algblocktwo{0.4}{0.98}{3.82}{5.17}{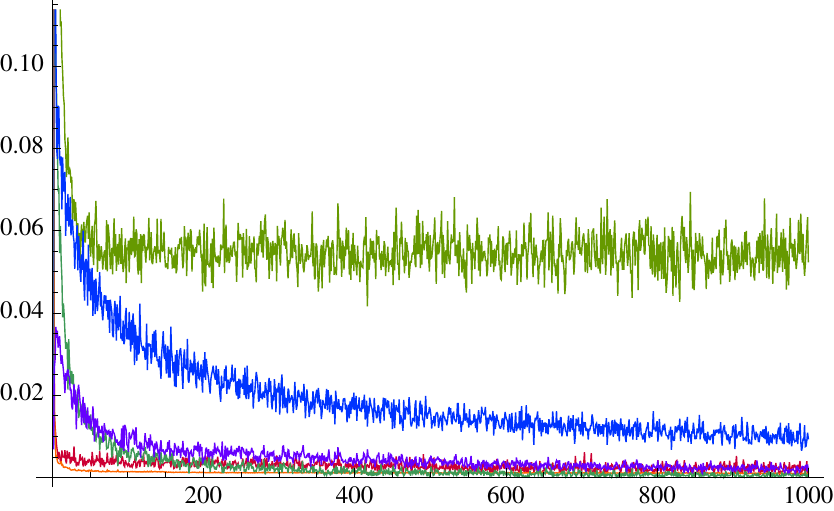}{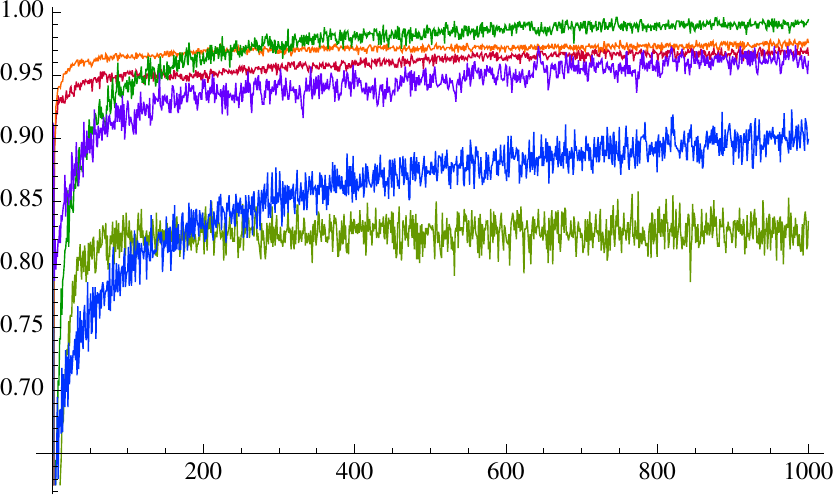}

\algblockone{1}{0.05}{32.8}{0.1}{52.2}{0.05}{65.1}{20.4}
\algblocktwo{0.4}{0.98}{28.1}{34.3}{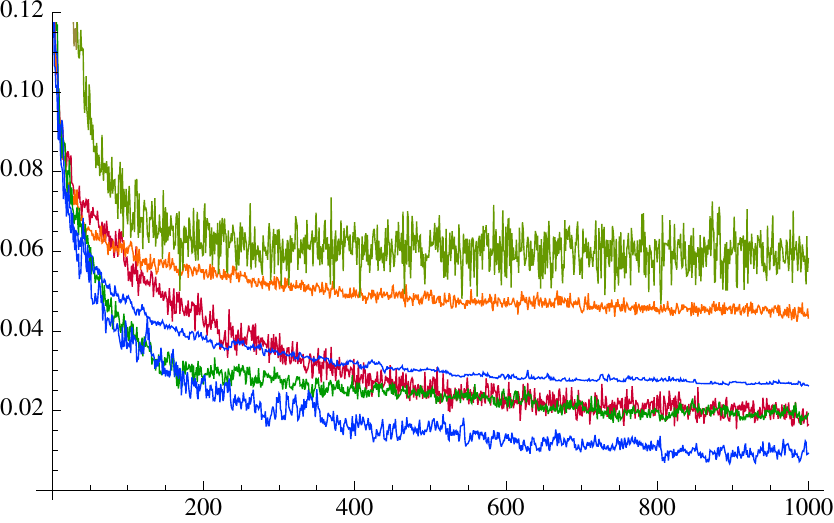}{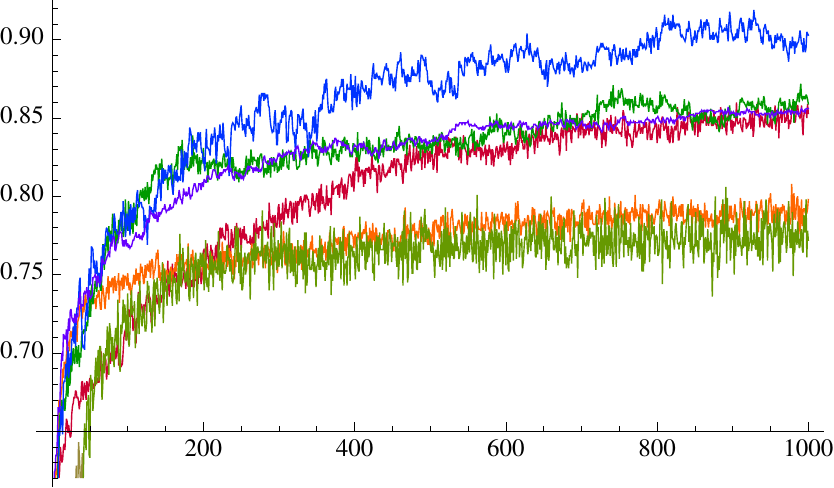}

\caption{Empirical Results for 2 arms, with different values of the variance}\label{fig:2arms}
\end{figure*}

\begin{figure*}[hp]

\algblockone{0.01}{0.005}{5.05}{0.001}{1.68}{0.4}{57.4}{67.2}
\algblocktwo{0.1}{0.95}{10.1}{17.3}{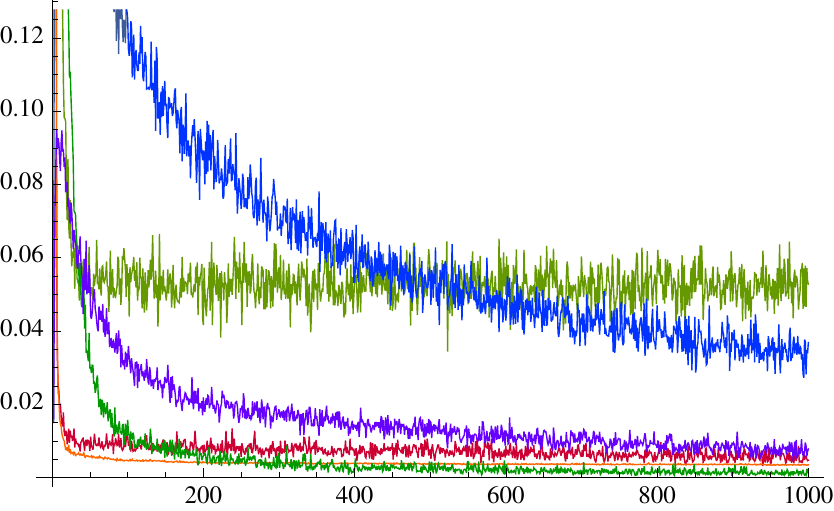}{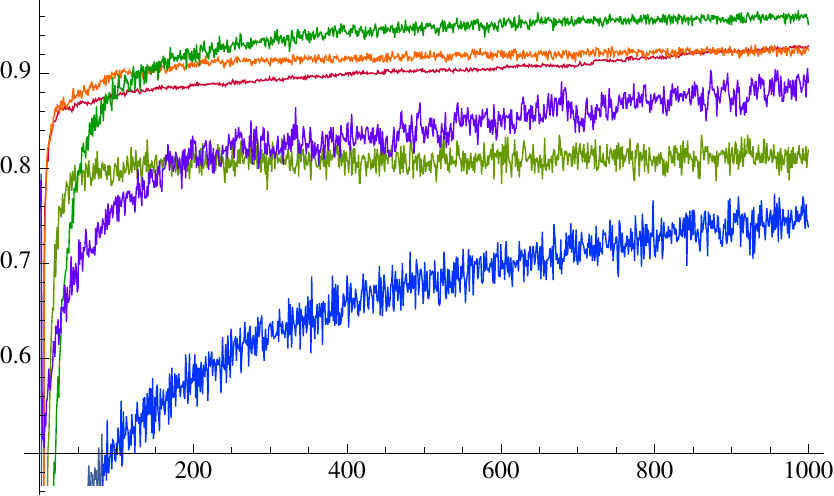}

\algblockone{0.1}{0.001}{8.93}{0.01}{5.63}{0.25}{55.8}{67.4}
\algblocktwo{0.1}{0.95}{10.3}{17.5}{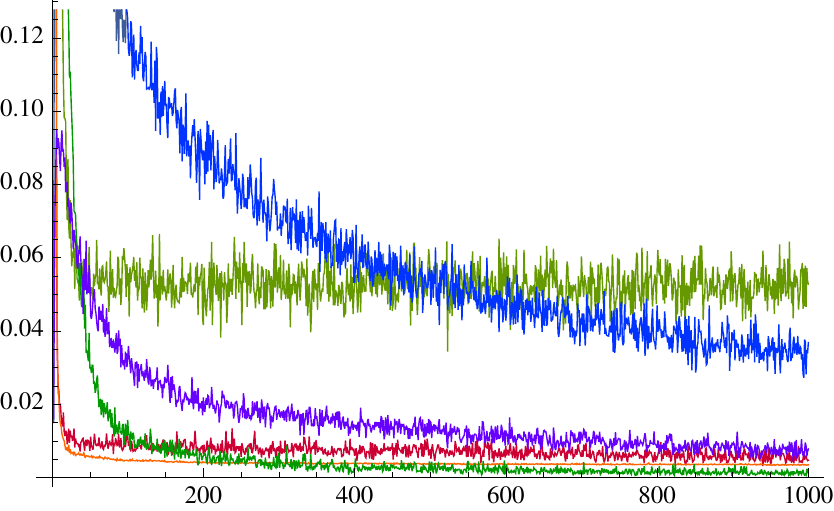}{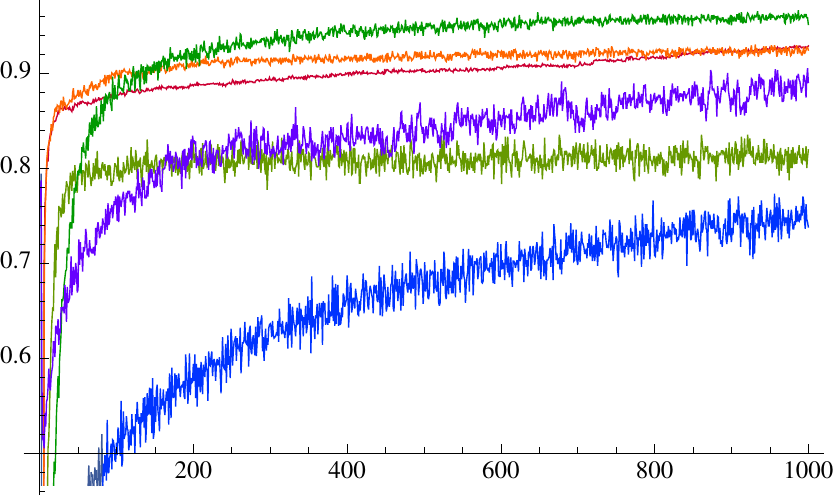}

\algblockone{1}{0.05}{71.192}{0.1}{77.4}{0.05}{106}{69.5}
\algblocktwo{0.01}{0.9}{46.9}{50.7}{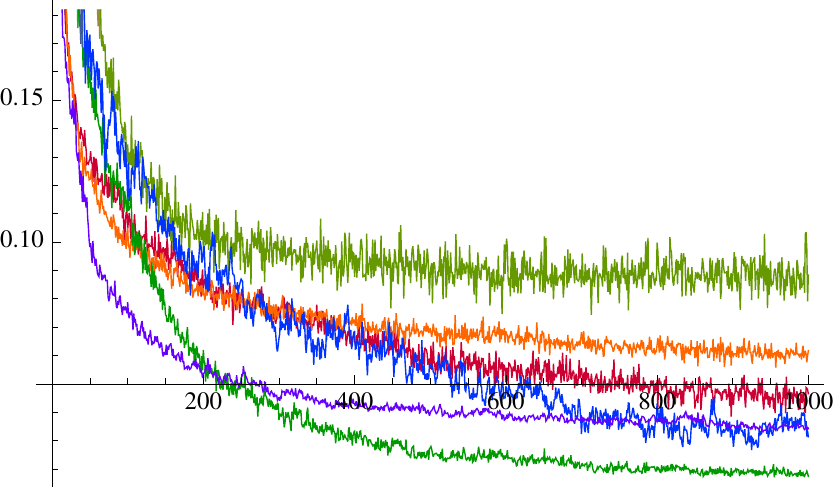}{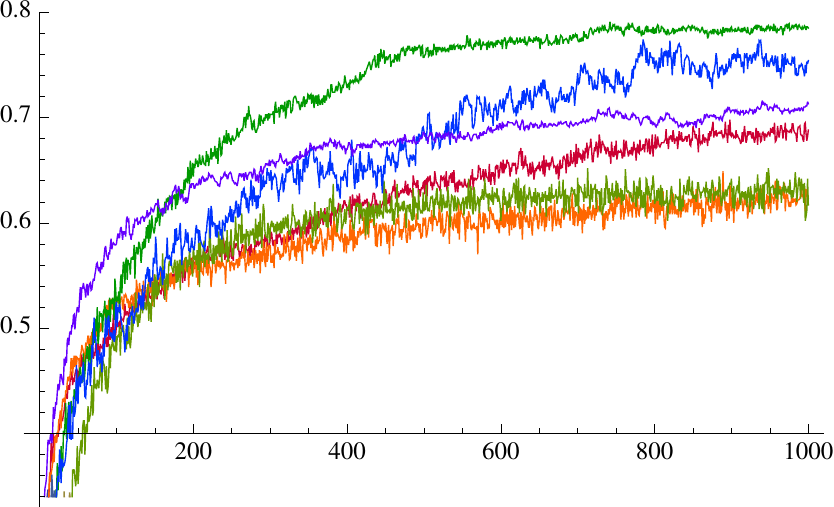}

\caption{Empirical Results for 5 arms, with different values of the variance}\label{fig:5arms}
\end{figure*}

\begin{figure*}[hp]

\algblockone{0.01}{0.001}{8.31}{0.001}{4.18}{0.5}{43.2}{127}
\algblocktwo{0.01}{0.98}{22.0}{34.7}{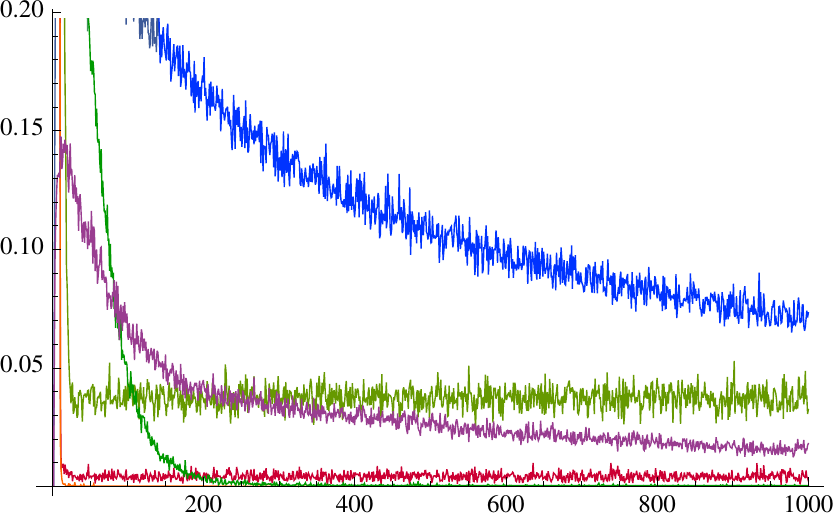}{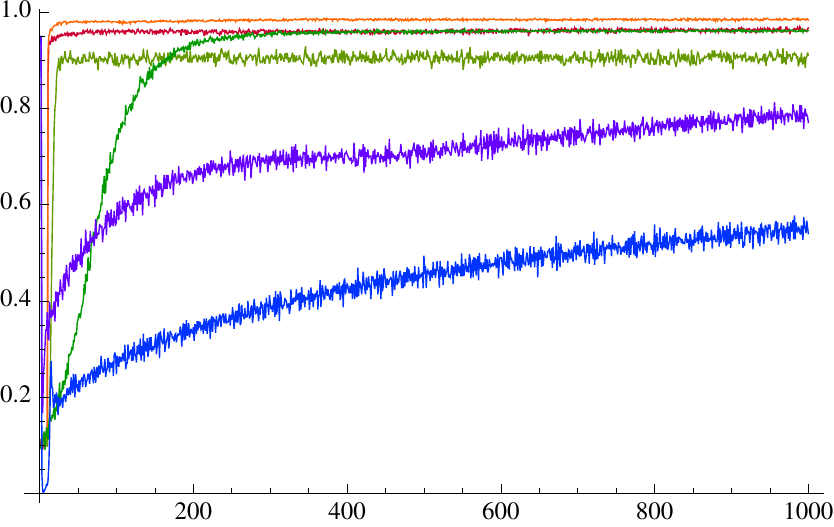}

\algblockone{0.1}{0.005}{12.9}{0.01}{8.24}{0.5}{48.6}{127}
\algblocktwo{0.1}{0.95}{23.3}{34.0}{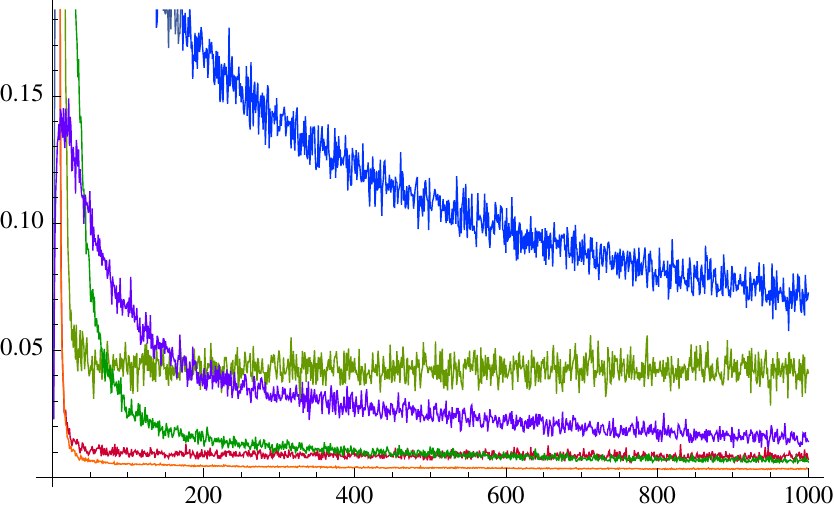}{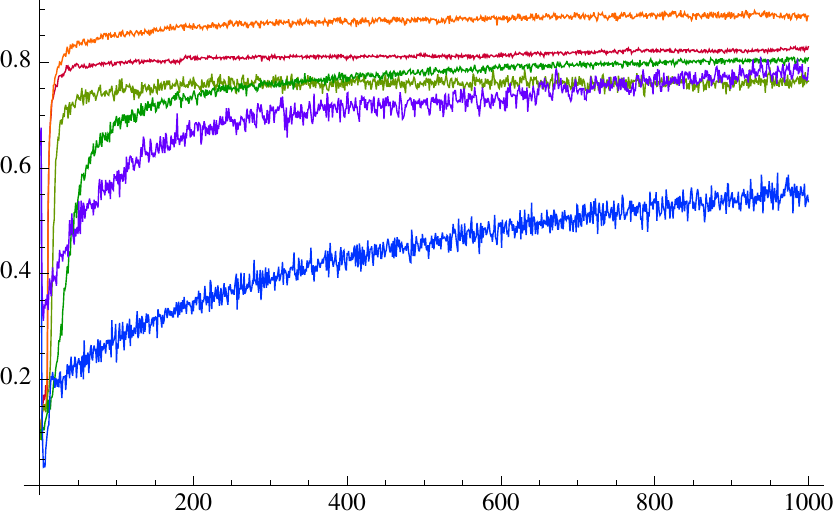}

\algblockone{1}{0.1}{102}{0.05}{87.9}{0.05}{110}{128}
\algblocktwo{0.1}{0.98}{91.1}{63.7}{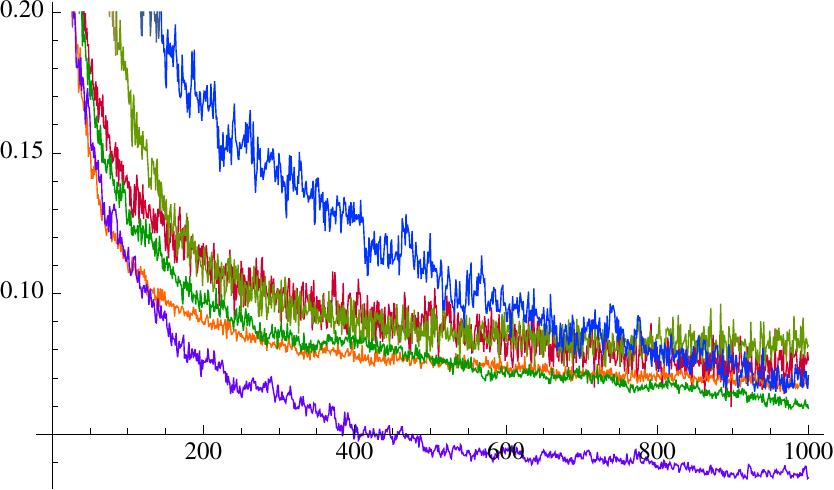}{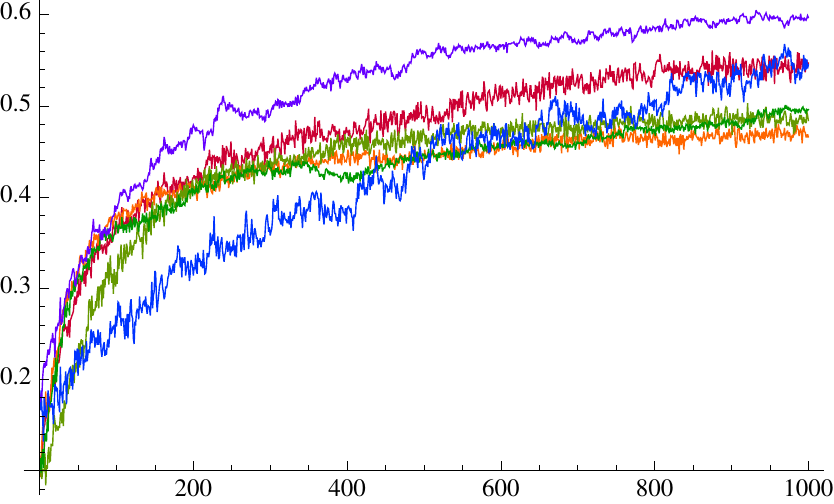}

\caption{Empirical Results for 10 arms, with different values of the variance}\label{fig:10arms}
\end{figure*}

\begin{figure*}[hp]

\algblockone{0.01}{0.005}{29.2}{0.001}{24.3}{0.5}{47.2}{296}
\algblocktwo{0.01}{0.98}{78.3}{119}{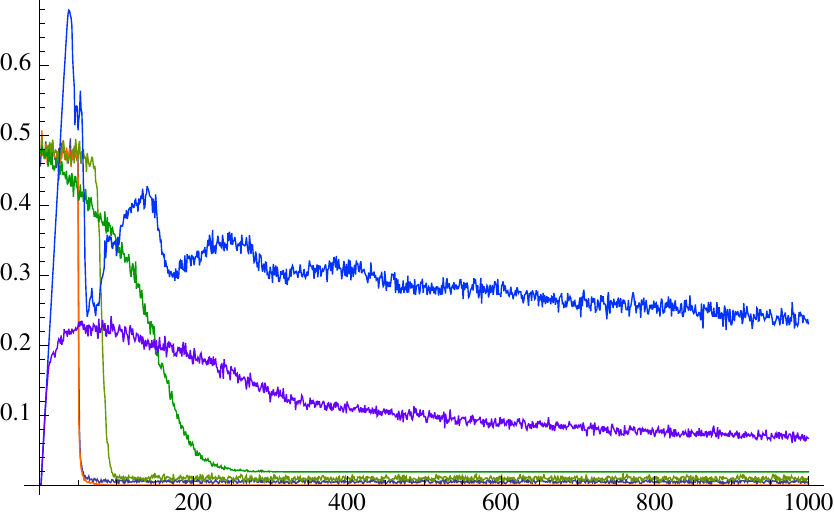}{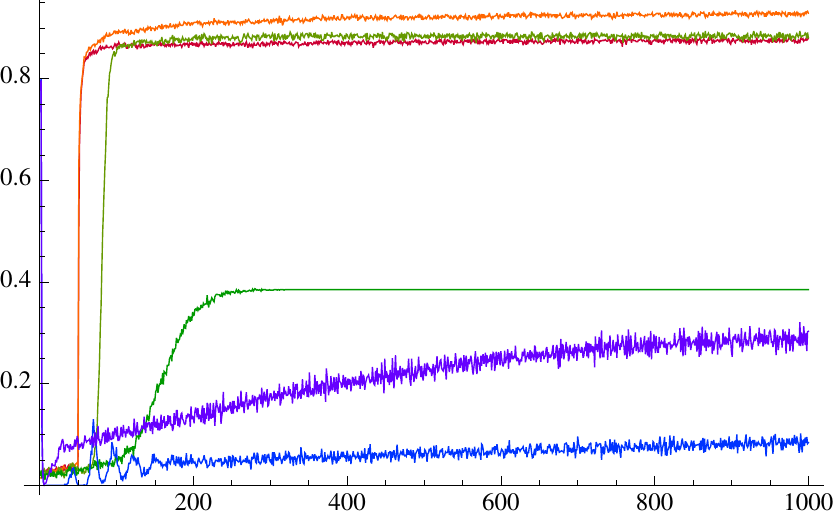}

\algblockone{0.1}{0.005}{39.5}{0.01}{32.0}{0.5}{56.7}{296}
\algblocktwo{0.1}{0.98}{80.0}{120}{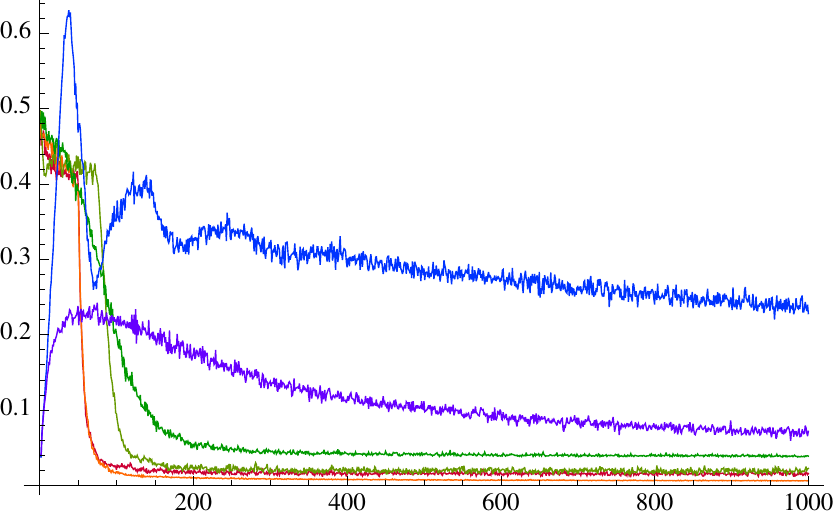}{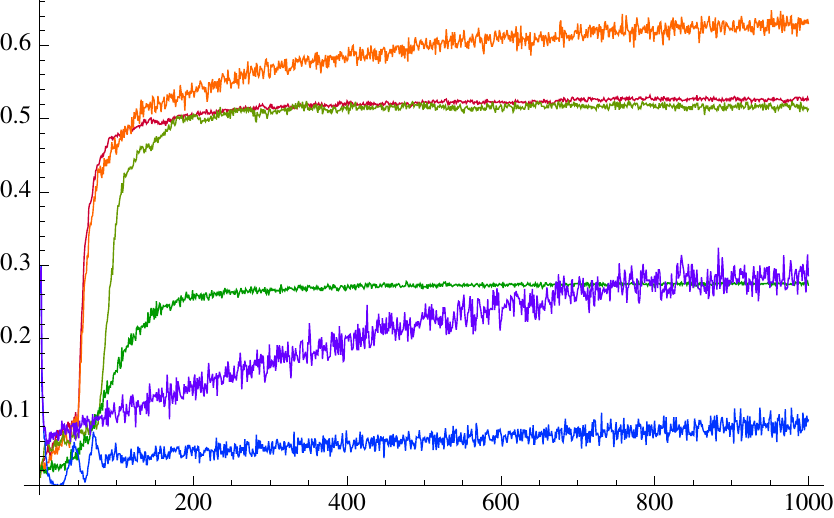}

\algblockone{1}{0.01}{112}{0.007}{107}{0.5}{130}{297}
\algblocktwo{0.01}{0.98}{123}{147}{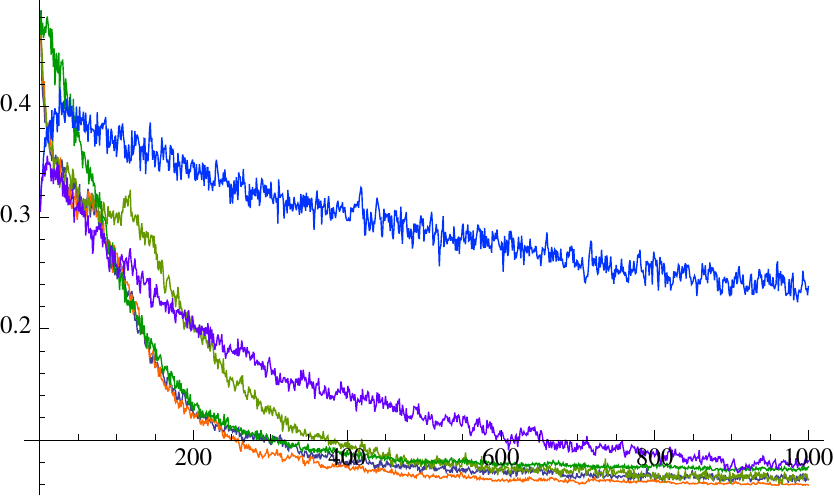}{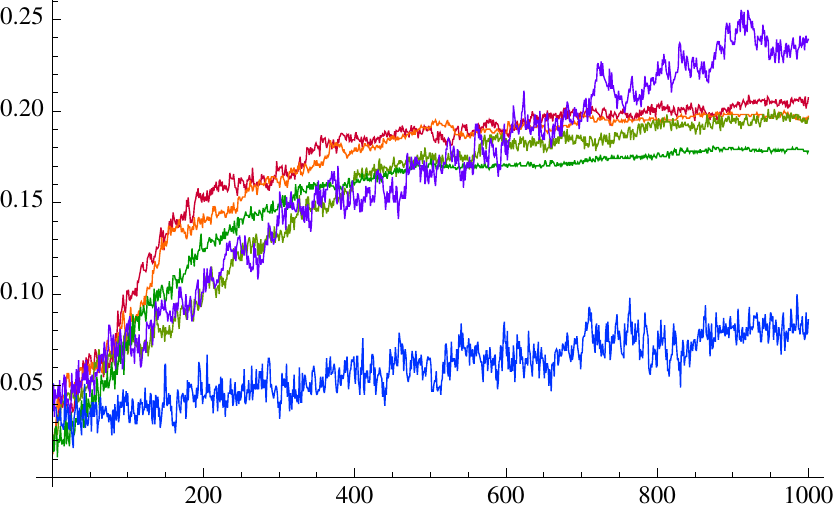}

\caption{Empirical Results for 50 arms, with different values of the variance}\label{fig:50arms}
\end{figure*}

Our main set of results is presented in Figure \ref{fig:2arms} for $K=2$, Figure \ref{fig:5arms} for $K=5$, Figure \ref{fig:10arms} for $K=10$ and Figure \ref{fig:50arms} for $K=50$.  In each case, we report for every value of the variance the total regret (as a numerical value) achieved by each algorithm. We also present graphs of the regret and of the percentage of optimal plays with respect to time.

The most striking observation is that the simplest algorithms, $\epsilon$-greedy and Boltzmann exploration, outperform their competitors on almost all tasks. Both heuristics perform very similarly, with softmax usually being slightly better. In particular, softmax outperforms the other algorithms in terms of total regret on all tasks, except on the high-variance setting ($\sigma=1$) for small and medium numbers of arms ($K=2, 5, 10$).  On these settings, softmax comes in second behind the UCB1-Tuned algorithm.  
In some sense, the fact that UCB1-Tuned was specifically designed to be sensitive to the variance of the arms justifies the fact that it should be superior at high values of the variance.  

There is relatively little to note about the performance of other algorithms. Pursuit methods perform overall the worst, because they plateau on sub-optimal solutions after $1000$ time steps. UCB1 converges to a solution much more slowly than the other algorithms, although that final solution appears to be very good, which is consistent with the results reported by Auer et al. \nocite{auer02}  Reinforcement comparison generates good average regret per turn towards the end of 1000 turns for small numbers of arms ($K=2, 5$), but starts trailing the other algorithms at larger values of $K$ ($K=10, 50$).  Its total regret is relatively high because it is slower in the beginning (due to the need to estimate the average reward). Overall, our results suggest that there is no advantage to using pursuit and reinforcement comparison in practice.

A second important observation is that algorithms are affected differently by variations in the characteristics of the bandit. UCB methods for example, handle bandits with small numbers of arms and high reward variances very well, but their performance deteriorates much more quickly than that of other algorithms when $K$ becomes large. 
We have not found any reference to this kind of behavior both in the empirical and in the theoretical literature on bandits.

\subsection*{Other characteristics of the bandit problem}

\begin{figure*}[ht]
\begin{tabular}{l}
	\subfigure[Uniform distribution]{\includegraphics[width=7cm]{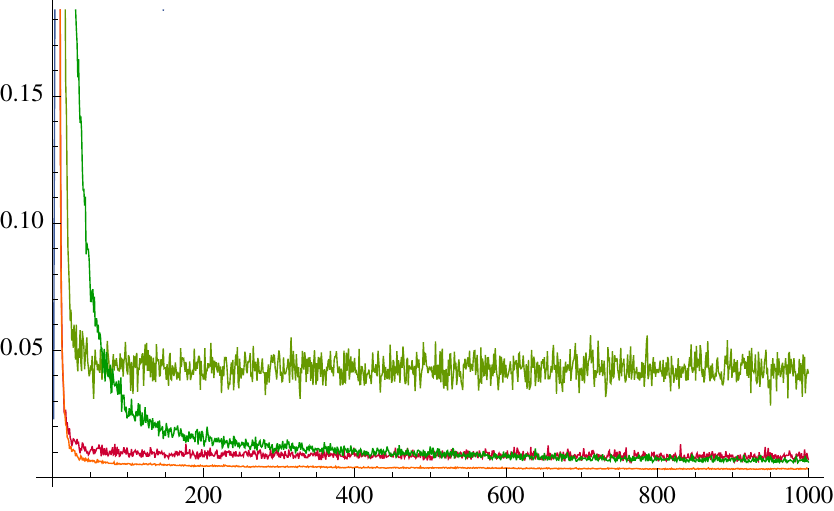}}
	\subfigure[Normal distribution]{\includegraphics[width=7cm]{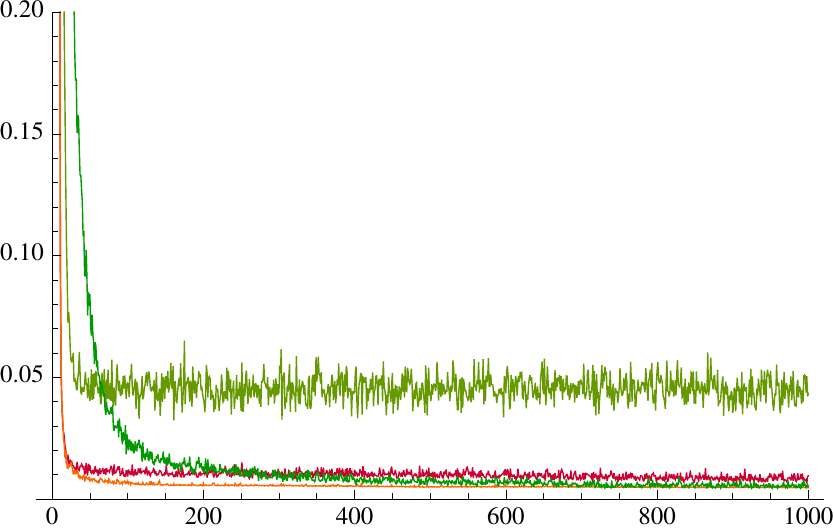}}\\
	\begin{tabular}{ll}
		\footnotesize \textcolor{red}{\rule{2mm}{2mm}} $\epsilon$-greedy, $\epsilon = 0.005$ &
		\footnotesize \textcolor{lightgreen}{\rule{2mm}{2mm}} Pursuit, $\beta = 0.5$  \\
		\footnotesize \textcolor{orange}{\rule{2mm}{2mm}} Softmax, $\tau = 0.01$ &
		\footnotesize \textcolor{green}{\rule{2mm}{2mm}} Reinforcement comparison, $\alpha = 0.1, \beta = 0.95$ \\
	\end{tabular}
\end{tabular} \\
\caption{Instantaneous regret per turn for various distributions of the expected values}\label{fig:mean_dist}
\end{figure*}

When the expected values for each reward distribution were sampled from a normal distribution, the algorithms performed slightly better than in the uniform setting. This is not surprising, as the best arm is better separated from the others when expected values are normally distributed. However, we did not observe any significant differences in the relative ranking of the algorithms, and we therefore omit detailed results. As an example of what we observed, we present in Figure \ref{fig:mean_dist} graphs for the setting $K=10$, $\sigma^2 = 0.1^2$.

\begin{figure*}[ht]
\subfigure[$\epsilon$-greedy]{\includegraphics[width=7cm]{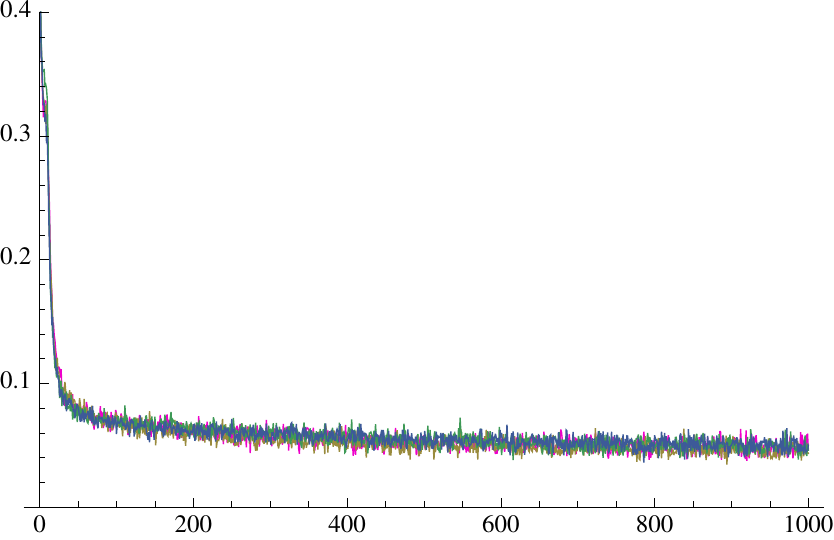}}
\subfigure[UCB1]{\includegraphics[width=7cm]{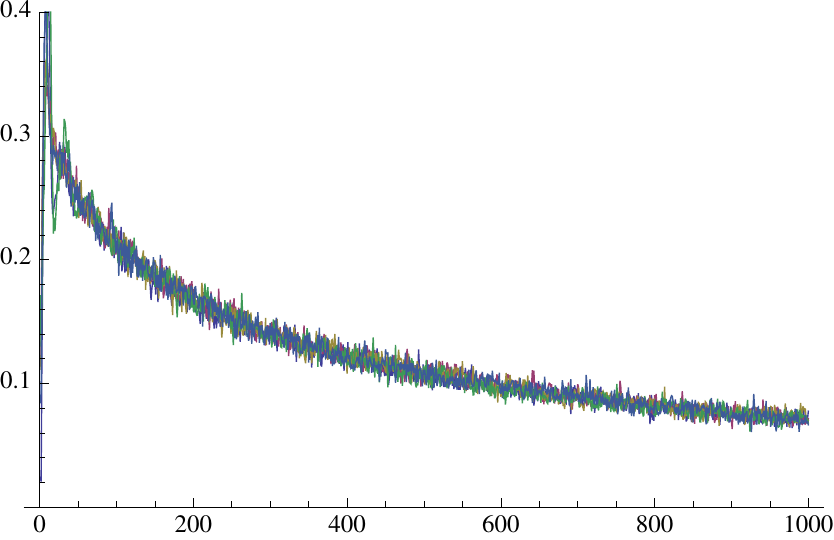}}
\caption{Average regret generated from the normal, triangular, uniform, inverse Gaussian, and Gumbel reward distributions. }\label{fig:dist}
\end{figure*}

Quite surprisingly, the type of reward distribution did not have any noticeable effect on the performance of the algorithms. See Figure \ref{fig:dist} for an example. We find this to be rather counter-intuitive, as one would expect, for example, that it would be harder to identify the best arm when its reward distribution is skewed to the left.

\subsection*{Parameter tuning}

We observed that algorithm parameters significantly affect performance. The average increase in total regret for an incorrectly tuned algorithm was roughly $20\%$, although in several cases the increase in regret was much larger. Surprisingly, a parameter value optimal on one bandit instance could suddenly become one of the worst after we had increased reward variance by only one notch. To illustrate this behavior, we show in Table \ref{tab:param} the total regret achieved by Boltzmann exploration given different values of $\tau$. Such measurements were carried out to tune every algorithm on every combination of $K$ and $\sigma^2$, and similar behavior was observed in every case.

\begin{table*}\begin{center}
\begin{tabular}{ccc}
\begin{tabular}{|c|c|}
\hline \multicolumn{2}{|c|}{\Large $\sigma = 0.01$}\\
\hline \textbf{Value of $\tau$} & \textbf{Total regret} \\
\hline 0.0007 & 4.29201 \\
\hline 0.001 & 4.18128 \\
\hline 0.005 & 4.24999 \\
\hline 0.01 & 4.84829 \\
\hline 0.05 & 19.8988 \\
\hline 0.1 & 54.1001 \\
\hline
\end{tabular} &
\begin{tabular}{|c|c|}
\hline
\multicolumn{2}{|c|}{\Large $\sigma = 0.1$}\\
\hline \textbf{Value of $\tau$} & \textbf{Total regret} \\
\hline 0.0007 & 12.5704 \\
\hline 0.001 & 11.3791 \\
\hline 0.005 & 11.0161 \\
\hline 0.01 & 8.23719 \\
\hline 0.05 & 19.6542 \\
\hline 0.1 & 53.4932 \\
\hline
\end{tabular} &
\begin{tabular}{|c|c|}
\hline
\multicolumn{2}{|c|}{\Large $\sigma = 1$}\\
\hline \textbf{Value of $\tau$} & \textbf{Total regret} \\
\hline 0.0007 & 118.749 \\
\hline 0.001 & 112.574 \\
\hline 0.005 & 110.298 \\
\hline 0.01 & 101.869 \\
\hline 0.05 & 87.9386 \\
\hline 0.1 & 95.3883 \\
\hline
\end{tabular} \\
\end{tabular}
\end{center}
\caption{The effects of parameters on Boltzmann exploration, on bandits with 10 arms and different reward variances.}\label{tab:param}
\end{table*}

Initially, we tuned the algorithms only once for each value of $K$, and used the same parameters for all values of $\sigma^2$. The results obtained were somewhat different from what we have reported above. We suspect that many empirical studies also fail to properly tune every algorithm, and many bandit strategies may be underperforming. The results we report in this section should therefore be carefully taken into account by authors doing further empirical studies.


\section{Discussion}\label{sec:disc}

Three important observations can be made from our results.

\subsection*{Simple heuristics outperform more advanced algorithms}

Admittedly, a similar observation has been made by Vermorel and Mohri (2005), but our experiments are the first to strongly suggest that this behavior occurs on practically every bandit problem instance. Our results also illustrate the magnitude of the advantage that simple heuristics offer. Boltzmann exploration generally outperforms its closest competitor by anywhere from $10\%$ to $100\%$. The advantage therefore appears to be quite substantial.

Our results also indicate the need for a formal analysis of simple heuristics, and more generally for the development of theoretically sound algorithms that perform as well as the simpler heuristics in practice. Even though Auer et al. (1998) derive polylogarithmic bounds for $\epsilon$-greedy and Boltzmann exploration, it remains an open problem to determine whether these simple heuristics are optimal, in the sense of achieving $O(\log T)$ regret. More generally, current theoretical bounds need to be improved to capture at least some of the rich behavior observed in experiments.

Since the problem of balancing exploration and exploitation appears throughout reinforcement learning, numerous algorithms employ bandit strategies as subroutines. For example, a well known algorithm that leverages bandit strategies for solving MDPs is UCT by Kocsis and Szepesvari (2004). The UCT algorithm, and many others, use the UCB1 bandit strategy because UCB1's theoretical properties can often be exploited in the analysis of the original algorithm. However, our results indicate that UCB-based algorithms should always be considered along ones based on $\epsilon$-greedy or softmax, since switching to these heuristics may yield substantial performance improvements in practice.

\subsection*{Algorithms' relative performance varies dramatically with the bandit instance}

Every algorithm has settings on which it performs well compared to other strategies and settings on which it performs poorly. In that sense, most algorithms possess specific strengths and weaknesses. The performance of the UCB family, for example, is excellent on bandits with a small number of arms and high reward variances, but degrades rapidly as the number of arms increases. These important algorithm properties are not described by any theoretical result, even though such a result would be valuable for choosing algorithms for specific problems.

In the absence of theoretical results, our empirical measurements can be used to guide the design of heuristics for specific bandit problems. As an example, consider a multi-stage clinical trial where the number and type of treatments varies significantly from stage to stage. The treatments' effectiveness is unknown, and we would like to both identify the best treatment and maximize the number of successfully treated patients. A good bandit-based heuristic for assigning patients to treatments should be based on a different algorithm at every stage, and our experiments suggest which ones to choose.

Finally, we would like to point out that the extreme variability of algorithm performance makes it necessary to evaluate algorithms on a wide range of settings, which past empirical studies have seldom done.

\subsection*{Only a subset of bandit characteristics are important for algorithm evaluation}

The relative performance of algorithms appears to be affected only by the number of arms and the reward variance. This has interesting implications for the type of regret bounds that we can expect to obtain. Recent theoretical work has focused on obtaining improved regret bounds by considering the reward variance of an arm (Audibert et al., 2009). Our results suggest that considering higher moments of the reward distribution will not be as fruitful, since the type of reward distribution had very little impact on algorithm performance.

More importantly, our experiments have precisely identified the aspects of a bandit problem that must be considered to accurately evaluate an algorithm, and we have made apparent the need to finely tune the algorithms on every bandit setting. Our experimental setup thus forms a good example of what is needed to accurately evaluate algorithms. If a similar methodology is adopted in subsequent studies, not only will algorithms be evaluated more accurately, but it will become easy to compare them between studies. In fact, we suggest that our measurements be used as a comparison point for other bandit strategies. Accurate and systematic measurements of algorithm performance would not only be very useful in practice, but could help direct research effort towards more promising algorithms. 

\section{Clinical trials}\label{sec:trials_intro}

We now turn our attention to an important application of bandit algorithms: the design of adaptive clinical trials. We will study whether bandit algorithms are well suited for allocating patients to treatments in a clinical trial.

In particular, we will be looking to answering the following three
questions:
\begin{enumerate}
\item Is it feasible to implement bandit strategies given real world constraints
such as patients' arrival dates, a long treatment time,
patient dropout, etc.?
\item At the end of the trial, can we identify the
best treatment with a good level of statistical confidence (with a
small p-value)?
\item Do bandit-based adaptive trials offer a significant advantage over traditional
trials in terms of patient welfare?
\end{enumerate}

To answer these questions, we simulate using real data what would have happened if a 2001-2002 clinical trial of opioid addiction treatments used adaptive bandit strategies instead of simple randomization. To measure the effectiveness of each approach, we use a variety of criteria, including
the number of patients successfully treated, patient retention, the number of adverse
effects, and others.

\section{Clinical context}\label{sec:trials_desc}

In this section, we describe the clinical trial on which we are basing our simulation.

The trial was conducted in 2001-2002 to compare the effectiveness of buprenorphine-naloxone (bupnal) and clonidine (clon) for treating opioid addiction, among in-patients and out-patients within community
treatment programs. Its primary goal was to confirm earlier results that established bupnal as the superior treatment; it was therefore a stage 4 trial. An adaptive strategy would have been particularly suited
in this context, since the identification of the best treatment has already
been done to a large extent.

Patients were randomized in a fixed 2:1 ratio, with
the majority being assigned to bupnal. Initially, 360 in-patients
and 360 out-patients were expected to be enrolled in the study. However,
the trial was terminated earlier than expected, and only 113 in-patients
(77 bupnal, 36 clon) and 231 out-patients (157 bupnal, 74 clon) participated.

Patients arrived over a period of 30 weeks, and were admitted into
the trial on each Wednesday. They received treatment for 13 days,
after which they were tested for opioids using a urine test. A patient
providing an opioid-free urine sample was considered to be successfully
treated, while an opioid-positive result or the failure to provide
a urine sample was considered a treatment failure.

Among in-patients, 59 of the 77 (77\%) individuals assigned to bupnal achieved
treatment success, compared to 8 of the 36 (22\%) clon patients. Among
out-patients, 46 of the 157 (29\%) bupnal individuals achieved the success
criterion, compared to 4 of the 74 (5\%) assigned to clonidine.

The patients' condition was also measured over the course
of the treatment. On every day, any adverse effects were recorded, 
and several tests designed to measure the patients'
level of well-being were administered.

In our simulation, we will use data from two such tests: the ARSW and the VAS. The
ARSW consists of a series of 16 observations performed by a doctor
and recorded as a numerical value on a scale from 1 to 9. For example, the doctor would estimate how irritated the patient is on that day and how watery their eyes appear.
The summary ARSW result
is defined as the sum of the 16 observations.
The VAS test consists in asking the patient to place a mark on a 100cm
line to indicate the strength of his craving for opiates, with 0cm
indicating no craving, and 100cm indicating extreme craving.

The raw data for this trial is publicly available on the Internet. Unfortunately, it is lacking detailed documentation, and it is unclear how it was interpreted to obtain
some of the results reported in Link (2005) \nocite{lin05}.
We therefore had to make certain minor assumptions regarding the data, but since they are the same for all algorithms, they do
not affect the validity of our results. We present these assumptions in Appendix \ref{app:preprocessing}.

\section{Experimental setup}\label{sec:trials_setup}

The simulation was implemented in Python and the source code of our program is available upon request. 
For each treatment strategy (the bandit algorithms and simple randomization), and each class of patients (the in-patients and the out-patients), $1000$ simulations of the trial were performed. All the results presented in this report form an average over these 1000 simulations.

Each simulation proceeds as follows. A total of 360 patients (representing the size of the population
expected to be enrolled in the original study) are assigned arrival
dates at random over a period of 30 weeks (the length of the original
trial). The treatment strategy then picks a treatment for each patient
in the order of their arrival into the study. Decisions are only
based on information available on the arrival day, and results are
fed back into the algorithm only two weeks later. In other words,
an algorithm can observe the outcome of assigning a patient to bupnal
on week 12 only on week 14. The outcome of that patient is not available
when processing patients coming in on weeks 12 and 13.


Once the treatment strategy assigns a patient to either bupnal or clon,
we determine the outcome, participation period, craving levels, adverse
effects, and all other results for that patient by sampling at random with
replacement from the chosen treatment's population. We are thus making
bootstrap estimates of the patients' true results. Doing so, we preserve natural
relationships among attributes: for example, we take into account the fact
that a patient who is a negative responder is also likely to have
high craving ratings.

The rewards fed to the bandit algorithms were 1 if the patient had a
positive response, and 0 if he or she had a negative response. Thus the
algorithms were essentially playing a multi-armed bandit with Bernoulli
arms. For algorithms that admit initial empirical means, we set these values to $1$.

\section{Simulation results}\label{sec:trials_results}

The performance of the six bandit algorithms was very similar, therefore we will only present
results for $\epsilon$-greedy, softmax, UCB1, and UCB-Tuned. Results for in-patients and out-patients also exhibited similar features, therefore we only include in-patient results. Out-patient material is available in Appendix \ref{app:out-patients}. The out-patient setting is in fact harder than the in-patient setting, as the success rate for clon among out-patients is only $5\%$, and a minimal number of clon successes is required to obtain a good confidence level on the best treatment.

\subsection*{Number of patients treated}

\begin{figure}[ht]
\begin{tabular}{cc}
	\subfigure[$\epsilon$-greedy]{\includegraphics[width=7cm]{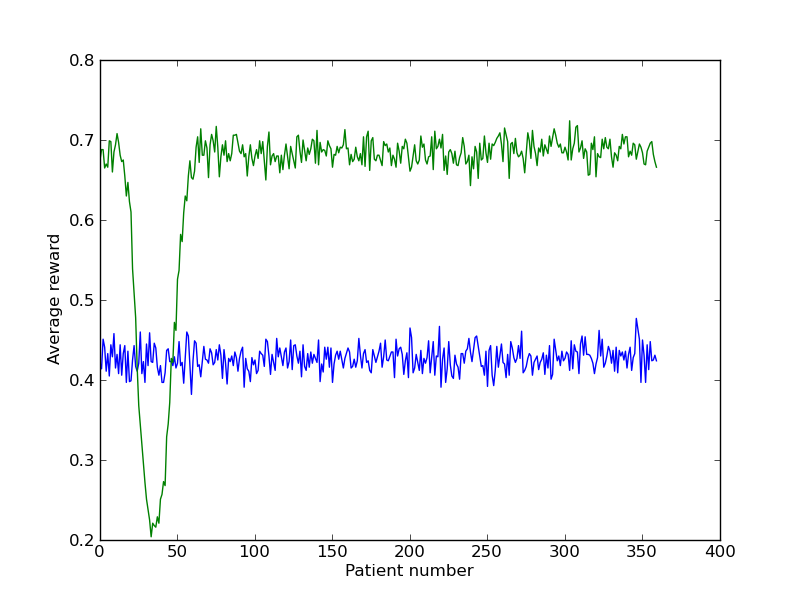}} &
	\subfigure[Softmax]{\includegraphics[width=7cm]{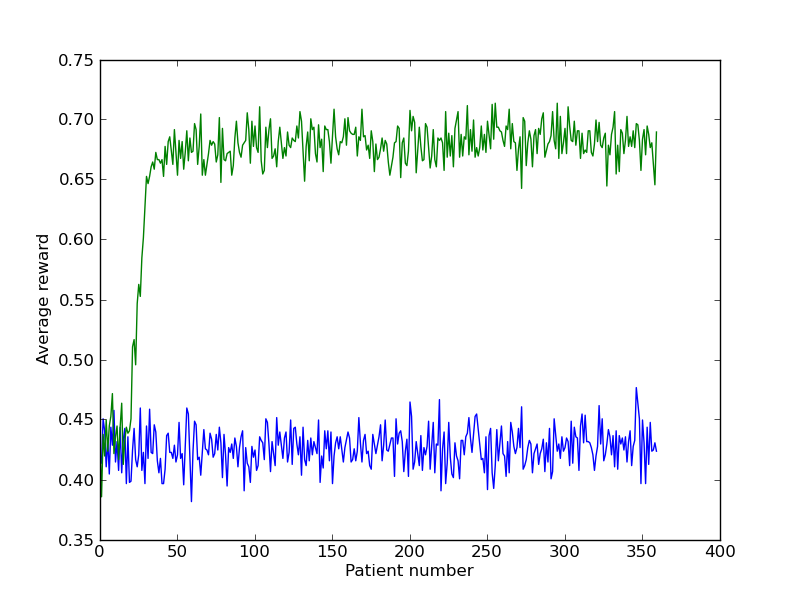}} \\
	\subfigure[UCB1]{\includegraphics[width=7cm]{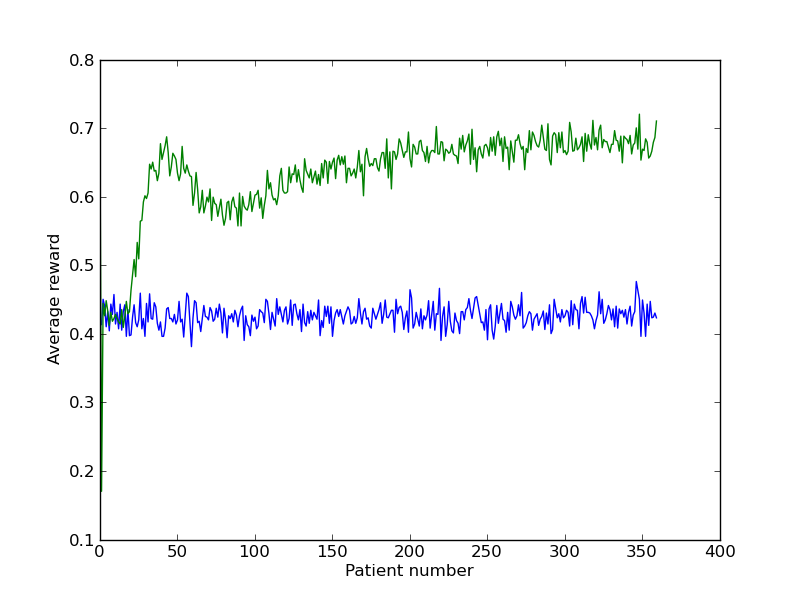}} &
	\subfigure[UCB1-Tuned]{\includegraphics[width=7cm]{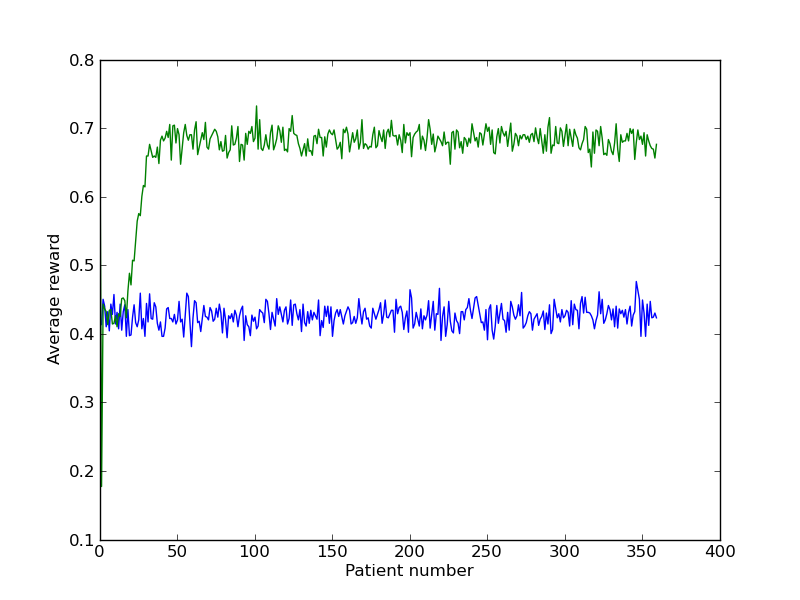}} \\
\end{tabular}
\caption{Average number of patients treated per day: Epsilon greedy (green) and randomization (blue) for in-patient
data}
\label{fig:patients_treated_per_day}
\end{figure}

In Figure \ref{fig:patients_treated_per_day}, we display for every algorithm the average number of patients treated per turn, with the average being taken over the $1000$ repetitions of the simulated trial. Figure \ref{fig:patients_treated_per_day} directly corresponds to the instantaneous regret plots from the first half of the paper.

We also report the average total number of patients treated over the duration of trial, with the average again being taken over the $1000$ simulated trial runs. These results
are shown in Table \ref{tab:number_of_patients_treated}.

\begin{table}[ht]
\begin{centering}
\begin{tabular}{|c|c|}
\hline 
Algorithm & Average number of patients treated\tabularnewline
\hline
\hline 
Randomization & 154.2\tabularnewline
\hline 
Epsilon Greedy & 235.6\tabularnewline
\hline 
Softmax & 239.2\tabularnewline
\hline 
UCB1 & 227.9\tabularnewline
\hline 
UCB-Tuned & 240.7\tabularnewline
\hline
\end{tabular}
\par\end{centering}

\caption{Number of patients treated}
\label{tab:number_of_patients_treated}
\end{table}

\subsection*{P-values}

An adaptive clinical trial must maximize the number of treated patients, but more importantly it must also identify the best treatment with a high level of statistical confidence. Obtaining rigorous statistical guarantees on the performance of the best treatment in an adaptive trial is generally considered difficult and is often cited as one of the main causes of their limited adoption (see Chow and Chang, 2008\nocite{CC08}).

In this study, however, we take the approach of Yao and Wei (1996)\nocite{YW96}, and compute p-values for the null hypothesis that the two treatments have equal probabilites of success using a simple $\chi^{2}$ test. The $\chi^2$ test for independence is very common in the clinical trials literature (see Sokal and Rohlf, 1981\nocite{SR81}), and although it may give weaker guarantees than more advanced statistical techniques, the p-values it produces hold even in the adaptive trial setting.

Table \ref{tab:p-values} presents p-values and contingency tables for each treatment allocation strategy. 

\begin{table}[ht]
\begin{center}
\begin{tabular}{|c|c|c|}
\hline 
Algorithm & Contingency table & p-value\tabularnewline
\hline
\hline 
Randomization & \begin{tabular}{cccc}
 & Success & Failure & Total\tabularnewline
Bupnal & 124.0 & 56.0 & 180.0\tabularnewline
Clon & 30.2 & 149.8 & 180.0\tabularnewline
Total & 154.2 & 205.8 & 360.0\tabularnewline
\end{tabular} & $0.0$\tabularnewline
\hline 
Epsilon Greedy & \begin{tabular}{cccc}
 & Success & Failure & Total\tabularnewline
Bupnal & 231.6 & 104.6 & 336.2\tabularnewline
Clon & 4.0 & 19.8 & 23.8\tabularnewline
Total & 235.6 & 124.4 & 360.0\tabularnewline
\end{tabular} & $7.8\times10^{-7}$\tabularnewline
\hline 
Softmax & \begin{tabular}{cccc}
 & Success & Failure & Total\tabularnewline
Bupnal & 236.4 & 106.8 & 343.2\tabularnewline
Clon & 2.8 & 14 & 16.8\tabularnewline
Total & 239.2 & 120.8 & 360.0\tabularnewline
\end{tabular} & $3.1\times10^{-5}$\tabularnewline
\hline 
UCB1 & \begin{tabular}{cccc}
 & Success & Failure & Total\tabularnewline
Bupnal & 221.6 & 100.5 & 322.1\tabularnewline
Clon & 6.3 & 31.6 & 37.9\tabularnewline
Total & 227.9 & 132.1 & 360.0\tabularnewline
\end{tabular} & $2.9\times10^{-10}$\tabularnewline
\hline 
UCB-Tuned & \begin{tabular}{cccc}
 & Success & Failure & Total\tabularnewline
Bupnal & 238.3 & 107.3 & 345.6\tabularnewline
Clon & 2.5 & 11.9 & 14.4\tabularnewline
Total & 240.8 & 119.2 & 360.0\tabularnewline
\end{tabular} & $1.5\times10^{-4}$\tabularnewline
\hline
\end{tabular}
\end{center}
\caption{Repartition of in-patients among the arms and p-values}
\label{tab:p-values}

\end{table}

\subsection*{Patient retention}

It is important for a clinical trial to retain a large fraction
of its patients until the end of treatment. This way more patients
are adequately treated, and just as importantly the clinical data can then be accurately processed using statistical methods.

\begin{figure}[ht]
\begin{center}
\includegraphics[width=10cm]{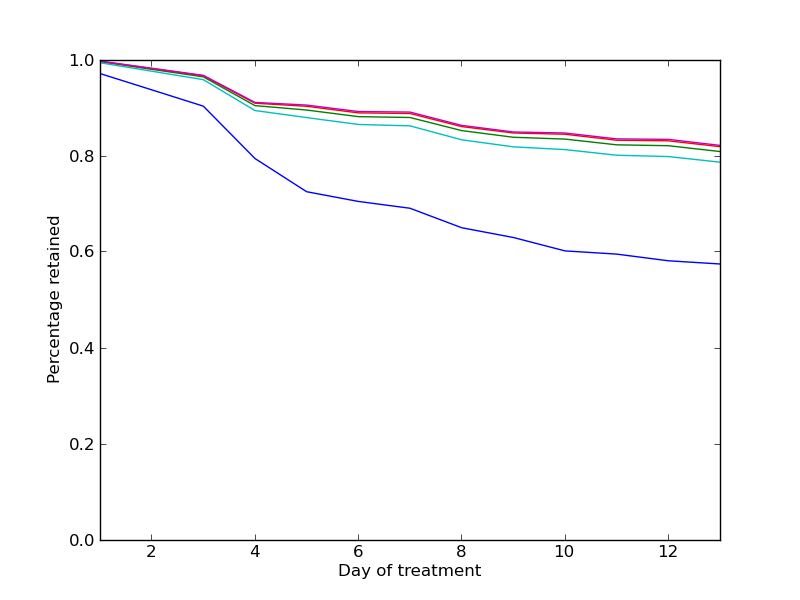}
\end{center}
\caption{Kaplan-Meier curves for in-patients. Simple randomization is in blue.}
\label{fig:kaplan-meier}
\end{figure}

Treatment retention is represented
using Kaplan-Meier curves, which show the percentage of patients remaining
in the treatment at each day. Figure \ref{fig:kaplan-meier} shows the Kaplan-Meier curves
for each algorithm. The bandit algorithms' curves are nearly indistinguishable from each other, hence we do not bother to annotate them.

\subsection*{Patient well-being}

\begin{figure}[ht]
\begin{center}
\includegraphics[width=10cm]{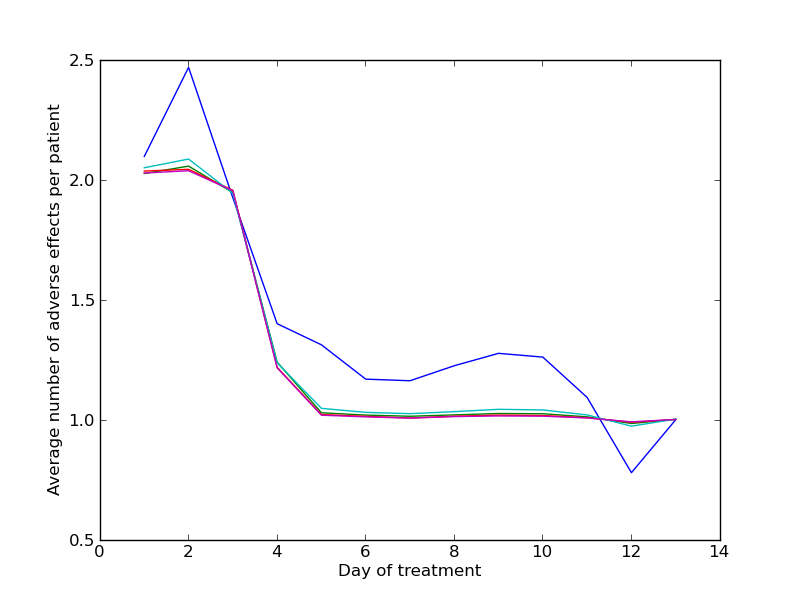}
\end{center}

\caption{Average number of adverse effects per day for in-patients. Simple randomization
is in blue.}
\label{fig:adverse_effects}
\end{figure}

In addition to the number of treated patients, we will be looking at the level of well-being of the patients while they are still in the treatment.
We measure patient well-being through the
number of adverse effects that occur. Figure \ref{fig:adverse_effects} shows the average
number of adverse effects per patient for each day of the treatment.
This value is computed by dividing the total number
of adverse effects on a given day by the number of patients still in
the treatment. Again, the bandit strategies are nearly indistinguishable from each other.

Finally, we consider the patients' craving ratings as a measure of
their well-being. Table \ref{tab:cravings} shows the mean ARSW and VAS test results
over all the patients.

\begin{table}[ht]
\begin{centering}
\begin{tabular}{|c|c|c|}
\hline 
Algorithm & ARSW & VAS\tabularnewline
\hline
\hline 
Randomization & 35.94 & 38.94\tabularnewline
\hline 
Epsilon greedy & 22.12 & 28.4\tabularnewline
\hline 
Softmax & 21.53 & 27.98\tabularnewline
\hline 
UCB1 & 23.32 & 29.4\tabularnewline
\hline 
UCB-Tuned & 21.29 & 27.76\tabularnewline
\hline
\end{tabular}
\par\end{centering}

\caption{Mean craving ratings for in-patients}
\label{tab:cravings}
\end{table}

\section{Discussion}\label{sec:trials_disc}

Overall, our results provide an affirmative answer to the three questions
we were looking to answer.

\subsection*{Is it practical to implement bandit algorithms?}

Even though the bandit algorithms received delayed feedback, the impact on their effectiveness was minimal. A total of 360 patients
were randomized over 30 weeks, with an average of 12 patients per
week. Thus on average, an algorithm had to make 24 decisions before
observing their outcomes. This represents about 7\% of the size
of the entire population and is clearly non-negligible.

Other practical constraints such as unknown arrival times and patient dropout
did not pose a problem in the context of this simulation. Since dropout
was interpreted as treatment failure, we were not required to fill in
missing treatment outcomes. This treatment of patient dropout
is inspired by the methodology of the initial study of Link (2005) \nocite{lin05}, and can be used in many clinical trials.

\subsection*{Can the algorithms identify the best treatment with high confidence?}

In almost every case, the null hypothesis that both treatments have
equal effectiveness could be rejected at a very high p-value. In the in-patient
setting, the worst p-value was $1.5\times10^{-4}$. Identifying the best treatment in the out-patient setting was much more difficult, as the success rate for clonidine was only 5\%, and the $\chi^2$ test requires a minimum number of successes for every treatment. As expected, bandit algorithms had more difficulty in obtaining a good p-value, but most of them still performed very well. The worst
p-value of $0.017$ was returned by Epsilon greedy. Although this
is much weaker than what simple randomization returned, even this confidence level
would generally be sufficient to establish the superiority of bupnal.
Other treatment strategies achieved much better p-values.
Therefore we conclude that bandit algorithms can reliably identify
the best arm. 

\subsection*{Are there medical advantages to using bandit-based adaptive strategies?}

Clearly, adaptive bandit-based clinical trials result in a large increase in patient
welfare. In both the in-patient and out-patient cases, at least 50\%
more patients were successfully treated. It is interesting to observe
that almost all the algorithms were able to find the best treatment
after processing about 50 patients, and from that point they administered
only the better treatment.

In addition, the percentage of patients still in the trial
after 13 days increased by almost 20\%, and much fewer
adverse effects were observed. In the out-patient setting, almost no adverse effects
occurred after day 3. The patients' levels of well-being were clearly better
in the bandit-based trials. Both the ARSW and the VAS scores
were almost 50\% lower with adaptive trials in the in-patient case.
These results strongly suggest that the average patient in
an adaptive trial was able to overcome their addiction more much easily.

\section{Conclusion}\label{sec:concl}

In this paper, we have presented an empirical study of the most popular algorithms for the multi-armed bandit problem. Most current theoretical guarantees do not accurately represent the real-world performance of bandit algorithms. Empirical results, on the other hand, usually apply to only a small number of strategies on a limited number of settings and fail to consider every important aspect of the bandit problem, most notably the reward variance.

To address these issues, we conducted our own empirical study. Our goals were to first identify the aspects of a bandit that affect the regret of a policy, and then to measure precisely how these aspects affect the regret. Surprisingly, we discovered that only the number of arms and the reward variance influence the performance of algorithms relative to each other.

We then proceeded to evaluate the performance of each algorithm under twelve values of these two parameters. The twelve settings were meant to cover most types of bandits that appear in practice. Surprisingly, simple heuristics consistently outperformed more advanced algorithms with strong theoretical guarantees. On almost every bandit problem instance, the softmax algorithm generated at least 50\% less regret than UCB1-Tuned, the best algorithm for which theoretical guarantees exist. This finding clearly shows there is a need for a theoretical analysis of simple heuristics.

In addition, we observed that algorithm performance varies significantly across bandit instances. We identified for each algorithm the instances on which it performs well, and the instances on which it performs poorly. This information can be exploited in practice when designing heuristics for real-world instances of the bandit problem.

Finally, our study precisely identified the aspects of a bandit problem that must be considered in an experiment and demonstrated the need to tune every algorithm on every instance of the problem. Our experimental setup forms a good reference for future experiments, and the data we obtained can serve as a point of comparison for other algorithms.

In addition to including more algorithms and considering different variances and arm numbers, our study could be improved by considering settings where reward variances are not identical. Certain algorithms, such as UCB1-Tuned, are specifically designed to take into account the variance of the arms, and may therefore have an advantage in such settings. 

In the second half of the paper, we turned our attention to an important application of the bandit problem: clinical trials. Although clinical trials have motivated theoretical research on multi-armed bandits since Robbins' original paper, bandit algorithms have never been evaluated as treatment allocation strategies in a clinical trial.

In our study, we simulated using real data what would have happened if a 2001-2002 clinical trial used bandit strategies instead of simple randomization. 
We found that  bandit-based strategies successfully treated at least 50\% more patients and resulted in fewer adverse effects, fewer cravings,
and greater patient retention. At the end of the trial,
the best treatment could still be identified with a high degree of
statistical confidence. Our results offer compelling reasons to use adaptive clinical
trials, in particular ones that are based on multi-armed bandit
algorithms, for establishing the efficacy of medical treatments.

Our study could be further improved by considering a larger dataset, and a richer clinical setting, such as that of a multi-stage trial. Indeed, our experimental setup was very simple, and in particular we were unable to identify any significant difference in performance among algorithms.

More generally, bandit algorithms can be applied to problems in other fields, such as online advertising or network routing. Although online advertising datasets may be difficult to obtain, applying bandit algorithms in that field would be particularly interesting, since advertising is problem of great practical significance and existing empirical results are rather limited. 

Overall, we find surprising that a problem as broadly applicable as the bandit problem is so scarcely studied from an applications point of view. We hope our empirical study will encourage more authors to apply bandit algorithms to interesting real-world problems.

\bibliography{jmlr}

\newpage
\appendix

\section {Data preprocessing} \label{app:preprocessing}

The raw data for this trial is publicly available on the Internet, organized into 26 Excel tables.
However, it is lacking detailed documentation, and it is not clear how the data was interpreted to obtain
the results reported in Link (2005) \nocite{lin05}.

We therefore had to make the following decisions on how to interpret
the raw data. Since they are the same for all algorithms, they do
not affect the validity of the results obtained.
\begin{enumerate}
\item Only patients who had either {}``bupnal'' or {}``clon'' specified
in the {}``treatment'' column in the {}``demographics'' table
were considered.
\item Treatment response was determined to be positive if a patient had
an entry in the {}``lab results'' table indicating an opioid-negative
urine test at day 13 or 14 (the later of the two that is available).
Response was determined to be negative otherwise.
\item As in the original paper of Link (2005)\nocite{lin05}, a patient's participation period was defined as the number of
days between the first received medication and the last one. The first
medication was always administered on day one. We assumed that the last medication
was administered on the last treatment day for which the patient had
an entry in the {}``exposure'' table with a numerical value in the
{}``dose'' column and a non-empty entry in both columns identifying
the treatment day.
\item The average number of adverse effects per patient for each day was
calculated by taking the sum of all the adverse effects for that day,
and dividing by the number of patients that were present (presence
on a given day is defined as in \#3 above). The average number of
adverse effects per patient per day is then defined as the average
of the above averages.
\item Craving ratings were measured using two tests: the ARSW and the VAS.
For each test, an overall score was computed for each patient by averaging
all available VAS and summary ARSW scores. The average score for the
trial was then computed by averaging the above averages over all the
patients.
\end{enumerate}
This interpretation yielded results that were almost always within a few
percent of the ones reported in the paper.

\newpage

\section {Out-patient results} \label{app:out-patients}

The out-patient case is harder than the in-patient case since the
success rate for clon is only 5\%, and the $\chi^{2}$ test requires
a certain number of successes for each treatment to produce a good
p-value.

The $\chi^{2}$ test for independence normally assumes
a minimum of five samples in each cell of the contingency table. For cases
in which this assumption is not met, there exists a standard statistical
tool called Yates' correction for continuity (Yates, 1934\nocite{yat34}). It adjusts
the usual statistic for the fact that the table values are small.
We will be applying it every time the usual $\chi^{2}$
assumption is not met, and we will include the uncorrected p-value
where it is relevant.

\subsection*{Number of patients treated}

We will again present only results for Epsilon greedy, softmax, UCB1,
and UCB-Tuned.
\begin{figure}[ht]
\begin{tabular}{cc}
	\subfigure[$\epsilon$-greedy]{\includegraphics[width=7cm]{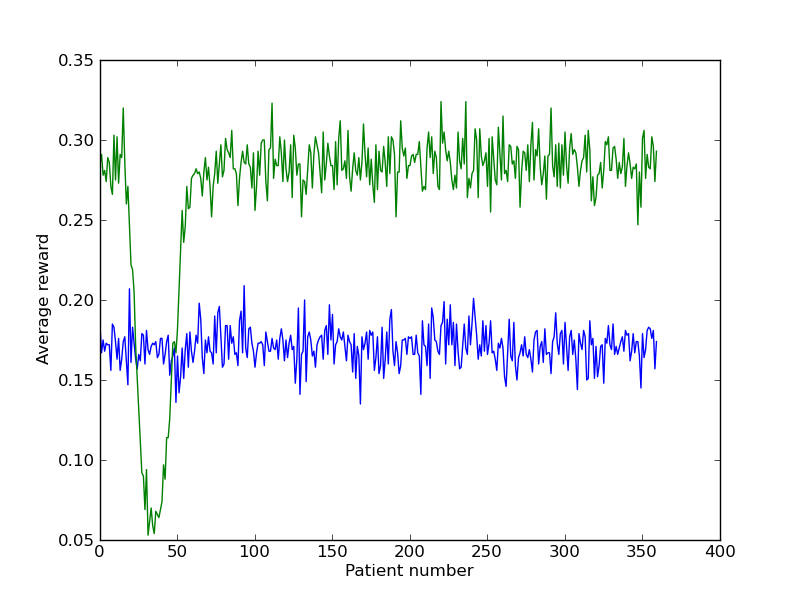}} &
	\subfigure[Softmax]{\includegraphics[width=7cm]{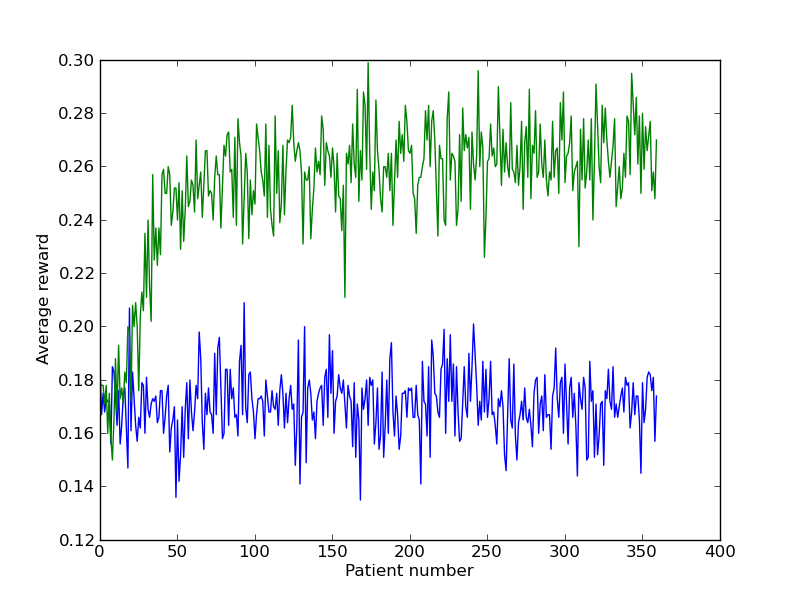}} \\
	\subfigure[UCB1]{\includegraphics[width=7cm]{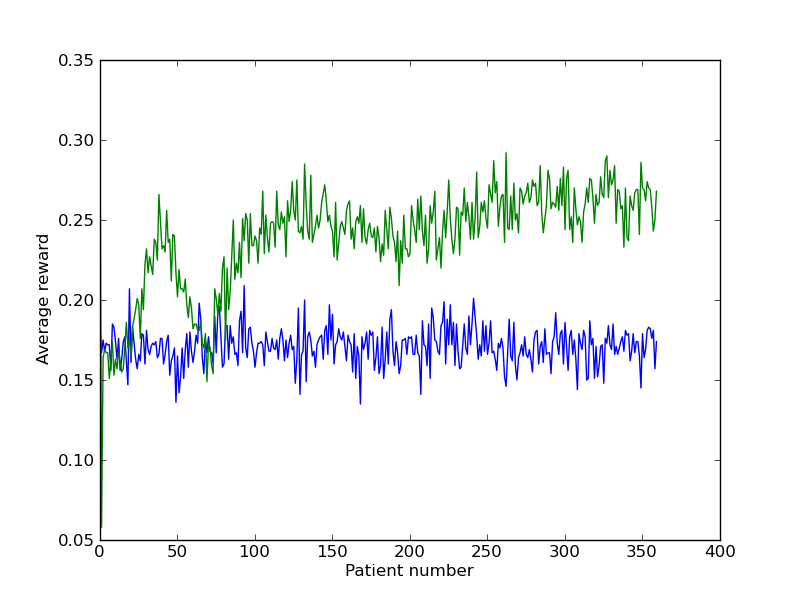}} &
	\subfigure[UCB1-Tuned]{\includegraphics[width=7cm]{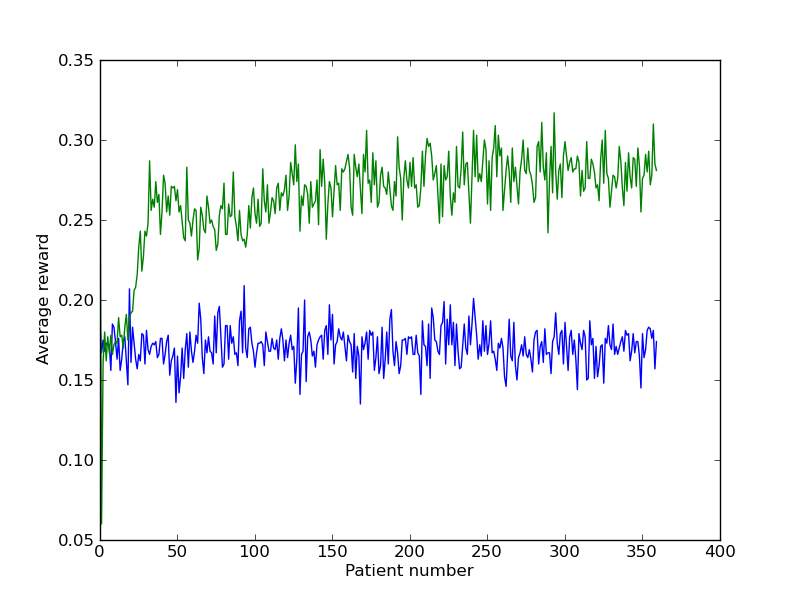}} \\
\end{tabular}
\caption{Average number of patients treated per day. Simple randomization is in blue}
\end{figure}

\begin{table}[ht]
\begin{centering}
\begin{tabular}{|c|c|}
\hline 
Algorithm & Average number of patients treated\tabularnewline
\hline
\hline 
Randomization & 61.8\tabularnewline
\hline 
Epsilon Greedy & 97.7\tabularnewline
\hline 
Softmax & 91.7\tabularnewline
\hline 
UCB1 & 86.3\tabularnewline
\hline 
UCB-Tuned & 95.8\tabularnewline
\hline
\end{tabular}
\par\end{centering}

\caption{Number of out-patients treated}

\end{table}

\subsection*{p-values}

We compute p-values for each algorithm like in the in-patient case.

\begin{table}[ht]
\begin{center}
\begin{tabular}{|c|c|c|}
\hline 
Algorithm & Contingency table & p-value\tabularnewline
\hline
\hline 
Randomizaiton & \begin{tabular}{cccc}
 & Success & Failure & Total\tabularnewline
Bupnal & 52.0 & 128.2 & 180.2\tabularnewline
Clon & 9.7 & 170.1 & 179.8\tabularnewline
Total & 61.7 & 298.3 & 360.0\tabularnewline
\end{tabular} & $3.5\times10^{-9}$\tabularnewline
\hline 
Epsilon Greedy & \begin{tabular}{cccc}
 & Success & Failure & Total\tabularnewline
Bupnal & 96.3 & 237.3 & 333.6\tabularnewline
Clon & 1.4 & 25 & 26.4\tabularnewline
Total & 97.7 & 262.3 & 360.0\tabularnewline
\end{tabular} & $0.017$ ($0.0088$ uncorr.)\tabularnewline
\hline 
Softmax & \begin{tabular}{cccc}
 & Success & Failure & Total\tabularnewline
Bupnal & 88.9 & 218.6 & 307.5\tabularnewline
Clon & 2.8 & 49.7 & 52.5\tabularnewline
Total & 91.7 & 268.3 & 360.0\tabularnewline
\end{tabular} & $5.6\times10^{-4}$\tabularnewline
\hline 
UCB1 & \begin{tabular}{cccc}
 & Success & Failure & Total\tabularnewline
Bupnal & 82.2 & 201.9 & 284.1\tabularnewline
Clon & 4.1 & 71.8 & 75.9\tabularnewline
Total & 86.3 & 273.7 & 360.0\tabularnewline
\end{tabular} & $3.9\times10^{-5}$\tabularnewline
\hline 
UCB-Tuned & \begin{tabular}{cccc}
 & Success & Failure & Total\tabularnewline
Bupnal & 93.9 & 230.7 & 324.6\tabularnewline
Clon & 1.9 & 33.5 & 35.4\tabularnewline
Total & 95.8 & 264.2 & 360.0\tabularnewline
\end{tabular} & $4.9\times10^{-3}$\tabularnewline
\hline
\end{tabular}

\caption{Repartition of out-patients among the arms and corresponding p-values}

\end{center}
\end{table}

\subsection*{Treatment retention}

\begin{figure}[ht]
\begin{center}
\includegraphics[width=10cm]{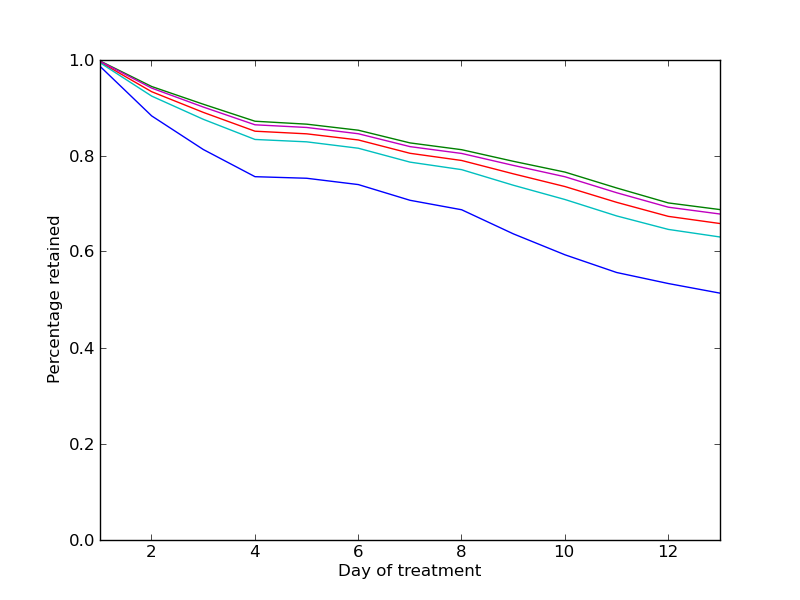}

\caption{Kaplan-Meier curves for out-patients. Simple randomization is in blue.} \label{fig:out_kaplan-meier}
\end{center}
\end{figure}

The Kaplan-Meier curves for out-patients are shown in Figure \ref{fig:out_kaplan-meier}.

\subsection*{Adverse effects}

\begin{figure}[ht]
\begin{center}
\includegraphics[width=10cm]{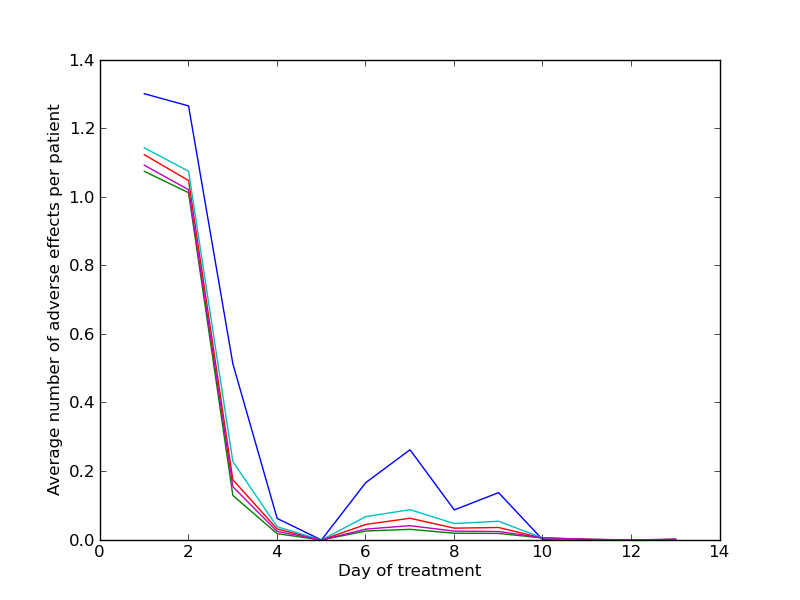}

\caption{Average adverse effects per day for out-patients. Simple randomization
is in blue.} \label{fig:out_adverse_effects}

\end{center}
\end{figure}

The adverse effects curves, defined like in the in-patient case, are
shown in Figure \ref{fig:out_adverse_effects}.

\subsection*{Craving ratings}

Table \ref{tab:out_cravings} shows the mean ARSW and VAS test results over all the out-patients.

\begin{table}[ht]
\begin{centering}
\begin{tabular}{|c|c|c|}
\hline 
Algorithm & ARSW & VAS\tabularnewline
\hline
\hline 
Randomization & 30.65 & 39.97\tabularnewline
\hline 
Epsilon greedy & 27.50 & 33.72\tabularnewline
\hline 
Softmax & 28.07 & 34.86\tabularnewline
\hline 
UCB1 & 28.45 & 35.79\tabularnewline
\hline 
UCB-Tuned & 27.66 & 34.2\tabularnewline
\hline
\end{tabular}
\par\end{centering}

\caption{Mean craving ratings for out-patients}
\label{tab:out_cravings}
\end{table}

\vskip 0.2in

\end{document}